\documentclass[11pt,letterpaper]{article}
\usepackage[letterpaper]{geometry}
\usepackage{acl2012}
\usepackage{times}
\usepackage{latexsym}
\usepackage[fleqn]{amsmath}
\usepackage{multirow}
\usepackage{url}
\makeatletter
\newcommand{\@BIBLABEL}{\@emptybiblabel}
\newcommand{\@emptybiblabel}[1]{}
\makeatother
\usepackage[
            pdfauthor={Edwin Simpson and Iryna Gurevych},
            pdftitle={Finding Convincing Arguments Using Scalable Bayesian Preference Learning},
            pdfsubject={Computer Science},
            pdfkeywords={argument mining, machine learning, NLP, Bayesian methods, variational inference, SVI, convincingness},
            hidelinks,
            bookmarks=false]{hyperref}

\usepackage[draft]{todonotes}

\usepackage{graphicx}
\usepackage{amsfonts}
\usepackage{algorithm2e}
\usepackage{array}
\usepackage[caption=false,font=footnotesize]{subfig}
\usepackage{url}
\usepackage{tabularx}
\usepackage{numprint}
\usepackage{multirow}

\newcommand{\bs}{\boldsymbol}


\title{ 
Finding Convincing Arguments Using Scalable Bayesian Preference Learning
}

\author{Edwin Simpson and Iryna Gurevych\\
  Ubiquitous Knowledge Processing Lab (UKP)\\
  Department of Computer Science \\
  Technische Universit{\"a}t Darmstadt  \\
  {\tt \{simpson, gurevych\}@ukp.informatik.tu-darmstadt.de} \\}

\begin{document}

\maketitle
\begin{abstract}
We introduce a scalable Bayesian preference learning method for identifying
convincing arguments in the absence of gold-standard ratings or rankings.
In contrast to previous work, we avoid the need for separate methods
to perform quality control on training data, predict rankings and perform pairwise classification.
Bayesian approaches are an effective solution when faced with sparse or noisy training data, 
but have not previously been used to identify convincing arguments.
One issue is scalability, which we address by developing a 
stochastic variational inference method for Gaussian process (GP) preference learning.
We show how our method can be applied to predict argument convincingness from crowdsourced data, 
outperforming the previous state-of-the-art, particularly when trained with small amounts of unreliable data.  
We demonstrate how the Bayesian approach enables more effective active learning,
thereby reducing the amount of data required to identify convincing arguments for new users and domains.
While word embeddings are principally used with neural networks, our results show that word embeddings in combination with linguistic features also benefit GPs when predicting argument convincingness.
\end{abstract}

%


\section{Introduction}\label{sec:intro}

Arguments are intended to persuade the audience of a particular point of view and 
are an important way for humans to reason about controversial topics~\cite{mercier2011humans}. 
The amount of argumentative text on any chosen subject can, however, overwhelm a reader.
Consider the scale of historical text archives 
or social media platforms with millions of users.
Automated methods could help readers overcome this challenge 
 by identifying high-quality, persuasive arguments from both sides of a debate. 
 
Theoretical approaches for assessing argument quality have proved difficult to apply to everyday arguments~\cite{boudry2015fake}.
Empirical machine learning approaches instead train models using example judgments of arguments,
such as those shown in Figure \ref{fig:argument_examples}.
\begin{figure}
\textbf{Topic:} ``William Farquhar ought to be honoured as the rightful founder of Singapore". \\
\textbf{Stance:} ``No, it is Raffles!" \\
\textbf{Argument 1:}  
HE HAS A BOSS(RAFFLES) HE HAS TO FOLLOW HIM AND NOT GO ABOUT DOING ANYTHING ELSE... \\
\textbf{Argument 2:} 
Raffles conceived a town plan to remodel Singapore into a modern city. The plan consisted of separate areas for different...\\
\textbf{Crowdsourced labels:} \{$2 \succ 1$, $1 \succ 2$, $2 \succ 1$\} 
\caption{Example argument pair from an online debate.}
\label{fig:argument_examples}
\end{figure}
Previous approaches to obtaining such judgments include training annotators to assign scores from 1-6~\cite{persing2017can}, 
asking annotators for simple binary or three-class categories~\cite{wei2016preliminary}, and
aggregating binary votes from multiple people~\cite{wei2016post,tan2016winning}.
However, these approaches are limited by the cost of training annotators, 
a highly restricted set of categories, or 
the need for multiple annotators per document.

An alternative way to judge arguments is to compare them against one another~\cite{habernal2016argument}.
When comparing the arguments in Figure \ref{fig:argument_examples}, we may judge that argument 1 is less convincing due to its writing style, whereas argument 2 
presents evidence in the form of historical events.
Pairwise comparisons such as this are known to place less cognitive burden on human annotators than  
choosing a numerical rating and allow fine-grained sorting of items that is not possible with categorical labels
~\cite{kendall1948rank,kingsley2006preference}.
Unlike numerical ratings, pairwise comparisons are not affected by different annotators' biases
toward high, low or middling values, or an individual's bias changing over time.

In practice, we face a data acquisition bottleneck when encountering new domains or audiences.
For example, neural network methods typically require datasets with 
many thousands of hand-labeled examples to perform well~\cite{srivastava2014dropout,collobert2011natural}.
One solution is to employ multiple non-specialist annotators at low cost (\emph{crowdsourcing}), 
but this requires quality control techniques to account for errors.
Another source of data are the actions of users of a software application, which can be interpreted as pairwise judgments~\cite{joachims2002optimizing}. For example, when a user clicks on an argument in a list it can be interpreted
 as a preference for the selected argument over more highly-ranked arguments.
However, the resulting pairwise labels are likely to be a very noisy indication of preference.

In this paper, we develop a Bayesian approach to learn from noisy pairwise preferences
based on Gaussian process preference learning (GPPL)~\cite{chu2005preference}.
We model argument convincingness as a function of textual features, including word embeddings,
and develop an inference method for GPPL that scales to realistic dataset sizes using stochastic variational inference (SVI) ~\cite{hoffman2013stochastic}. 
Using datasets provided by Habernal and Gurevych~\shortcite{habernal2016argument},
we show that our method outperforms the previous state-of-the-art for ranking arguments by convincingness 
and identifying the most convincing argument in a pair. 
Further experiments show that our approach is particularly advantageous with small, noisy datasets, 
and in an active learning set-up.
Our software is publicly available\footnote{\url{https://github.com/ukplab/tacl2018-preference-convincing}}.

The rest of the paper is structured as follows.
Section \ref{sec:related} reviews related work on argumentation,
then Section \ref{sec:bayesian} motivates the use of Bayesian methods by discussing their successful applications in NLP.
In Section \ref{sec:pref_learning}, we review preference learning methods and then Section \ref{sec:model}
describes our scalable Gaussian process-based approach.
Section \ref{sec:expts} presents our evaluation, 
comparing our method to the state-of-the art and testing with noisy data and active learning.
Finally, we present conclusions and future work.

\section{Identifying Convincing Arguments}\label{sec:related}

Lukin et al.~\shortcite{lukin2017argument} demonstrated that an audience's personality and prior stance affect
an argument's persuasiveness, but they were unable to predict belief change to a high degree of accuracy.
Related work has shown how persuasiveness is also affected by the sequence of arguments in a discussion 
~\cite{tan2016winning,rosenfeld2016providing,monteserin2013reinforcement},
but this work focuses on predicting salience of an argument given the state of the debate,
 rather than the qualities of arguments.

Wachsmuth et al.~\shortcite{wachsmuth2017argumentation} recently showed
 that relative comparisons of argument convincingness correlate with theory-derived quality ratings.
Habernal and Gurevych~\shortcite{habernal2016argument} established datasets
containing crowdsourced pairwise judgments of convincingness for arguments taken from online discussions. 
Errors in the crowdsourced data were handled by determining gold labels using the MACE algorithm~\cite{hovy2013learning}.
The gold labels were then used to train SVM and bi-directional long short-term memory (BiLSTM) classifiers to predict pairwise
labels for new arguments. 
The gold labels were also used to construct a directed graph of convincingness, which was input to PageRank 
to produce scores for each argument. 
These scores were then used to train SVM and BiLSTM regression models.
A drawback of such pipeline approaches is that they are prone to error propagation~\cite{chen2016joint},
and consensus algorithms, such as MACE, require multiple crowdsourced labels for each argument pair, 
which increases annotation costs.

\section{Bayesian Methods for NLP}\label{sec:bayesian}

When faced with a lack of reliable annotated data, 
Bayesian approaches have a number of advantages.
Bayesian inference provides a mathematical framework for combining multiple observations
with prior information. 
Given a model, $M$, and observed data, $D$, we apply Bayes' rule
to obtain a \emph{posterior distribution} over $M$, which can be used to make predictions 
about unseen data or latent variables:
\begin{equation}
  P(M|D) = \frac{P(D|M)P(M)}{P(D)},
  \label{eq:bayesrule}
\end{equation}
where $P(D|M)$ is the likelihood of the data given $M$, and $P(M)$ is the model prior.
If the dataset is small, the posterior remains close to the prior, so the model 
does not assume extreme values given a small training sample.
Rather than learning a posterior, neural network training typically selects model parameters that maximize the likelihood, 
so they are more prone to overfitting with small datasets, which can reduce performance~\cite{xiong2011bayesian}.

Bayesian methods can be trained using unsupervised or semi-supervised learning
to take advantage of structure in unlabeled data when labeled data is in short supply.
Popular examples in NLP are
Latent Dirichlet Allocation (LDA)~\cite{blei2003latent}, which is used for topic modelling,
and its extension, the hierarchical Dirichlet process (HDP)~\cite{teh2005sharing}, which learns the number of topics rather than requiring it to be fixed a priori.
Semi-supervised Bayesian learning
has also been used to achieve state-of-the-art results for semantic role labelling~\cite{titov2012bayesian}.

We can combine independent pieces of weak evidence using Bayesian methods through the likelihood.
For instance, a Bayesian network can be used to infer attack relations between arguments by combining votes for acceptable arguments from different people~\cite{kido2017}.
Other Bayesian approaches combine crowdsourced annotations to train a sentiment classifier
without a separate quality control step~\cite{simpson2015language,felt2016semantic}.

Several successful Bayesian approaches in NLP make use of Gaussian processes (GPs), which are 
distributions over functions of input features. 
GPs are nonparametric, meaning they can model highly nonlinear functions by
allowing function complexity to grow with the amount of data~\cite{rasmussen_gaussian_2006}.
They account for model uncertainty when extrapolating from sparse training data
and can be incorporated into larger graphical models.
Example applications include analyzing the relationship between a user's impact on Twitter 
and the textual features of their tweets~\cite{lampos2014predicting}, 
predicting the level of emotion in text~\cite{beck2014joint},
and estimating the quality of machine translations given source and translated texts~\cite{cohn2013modelling}.

\section{Preference Learning}\label{sec:pref_learning}

Our aim is to develop a Bayesian method for identifying convincing arguments 
given their features, which can be trained on noisy pairwise labels.
Each label, $i \succ j$, states that an argument, $i$, is more convincing than argument, $j$. 
This learning task is a form of \emph{preference learning}, which can be addressed in several ways.
A simple approach is to use a generic classifier by
obtaining a single feature vector for each pair in the training and test datasets,
either by concatenating the feature vectors of the items in the pair, 
or by computing the difference of the two feature vectors, as in SVM-Rank~\cite{joachims2002optimizing}. 
However, this approach does not produce ranked lists of convincing arguments without predicting a large number of pairwise labels, nor give scores of convincingness. 

Alternatively, we could learn an ordering over arguments directly using Mallows models~\cite{mallows1957non},
which define distributions over permutations.  
Mallows models can be trained from pairwise preferences 
~\cite{lu2011learning}, but inference is usually costly
since the number of possible permutations is $\mathcal{O}(N!)$, 
where $N$ is the number of arguments. 
Modeling only the ordering does not allow us to quantify 
the difference between arguments at similar ranks.

To avoid the problems of classifier-based and permutation-based methods, 
we propose to learn a real-valued convincingness function, $f$, that takes argument features as input
and can be used to predict rankings, pairwise labels, or ratings for individual arguments.
There are two well established approaches for mapping pairwise labels to real-valued scores: 
the Bradley-Terry-Plackett-Luce model~\cite{bradley1952rank,luce1959possible,plackett1975analysis}
and the Thurstone-Mosteller model~\cite{thurstone1927law,mosteller2006remarks}.
Based on the latter approach, 
Chu and Ghahramani~\shortcite{chu2005preference} introduced 
Gaussian process preference learning (GPPL), 
a Bayesian model that can tolerate errors in pairwise training labels
and gains the advantages of a GP for learning nonlinear functions from sparse datasets.
However, the inference method proposed by Chu and Ghahramani~\shortcite{chu2005preference} 
has memory and computational costs that scale with $\mathcal{O}(N^3)$,
making it unsuitable for real-world text datasets. 
The next section explains how we use recent developments in inference methods 
to develop scalable Bayesian preference learning for argument convincingness.
\section{Scalable Bayesian Preference Learning}\label{sec:model}

First, we introduce a probabilistic model for preference learning~\cite{chu2005preference}.
We observe preference pairs, each consisting of a pair of feature vectors $\mathbf x_i$ and $\mathbf x_j$, for arguments $i$ and $j$,
and a label $y \in \{i \succ j, j \succ i\}$.
We assume that the likelihood of $y$ depends on the latent convincingness, $f(\mathbf{x_i})$ and 
$f(\mathbf x_j)$, of the arguments in the pair. 
Our goal is to predict $y$ for pairs that have not been observed, 
and predict $f(\mathbf x_i)$, which may be used to rank arguments.
The relationship between convincingness and pairwise labels is described by the following:
\begin{flalign}
& p( i \succ j | f(\mathbf x_i), f(\mathbf x_j), \delta_{i}, \delta_{j} ) & \nonumber\\
& \hspace{0.9cm} = \begin{cases}
 1 & \text{if }f(\mathbf x_i) + \delta_{i} \geq f(j) + \delta_{j} \\
 0 & \text{otherwise,}
 \end{cases} &
 \label{eq:pl}
\end{flalign}
where $\delta \sim \mathcal{N}(0, 1)$ is Gaussian-distributed noise. 
If the convincingness $f(\mathbf x_i)$ is higher than the convincingness $f(\mathbf x_j)$, 
the preference label $i \succ j$ is more likely to be true.
However, the label also depends on the noise terms, $\delta_{i}$ and $\delta_{j}$,
to allow for errors caused by, for example, disagreement between human annotators.
We simplify Equation \ref{eq:pl} by integrating out $\delta_{i}$ and $\delta_{j}$ to obtain the \emph{preference likelihood}:
\begin{flalign}
& p( i \succ j | f(\mathbf x_i), f(\mathbf x_j) ) & \nonumber\\
& = \int\int p( i \succ j | f(\mathbf x_i), f(\mathbf x_j), \delta_{i}, \delta_{j} ) &\nonumber\\
& \hspace{3cm}\mathcal{N}(\delta_{i}; 0, 1)\mathcal{N}(\delta_{j}; 0, 1) d\delta_{i} d\delta_{j} &\nonumber\\
& = \Phi\left( z \right), 
\label{eq:plphi}
\end{flalign}
where $z = (f(\mathbf x_i) - f(\mathbf x_j)) / \sqrt{2}$,
and $\Phi$ is the cumulative distribution function of the standard normal distribution. 

We assume that convincingness is a function, $f$, of argument features, 
drawn from a Gaussian process prior: $f \sim \mathcal{GP}(0, k_{\theta}s)$, where 
$k_{\theta}$ is a kernel function with hyper-parameters $\theta$, 
and $s$ is a scale parameter. 
The kernel function controls the smoothness of $f$ over the feature space,
while $s$ controls the variance of $f$. 
Increasing $s$ means that, on average, the magnitude of $f(\mathbf x_i)-f(\mathbf x_j)$ increases  
so that $\Phi(z)$ is closer to $0$ or $1$, and erroneous pairwise labels are less likely.
Therefore, larger values of $s$ correspond to less observation noise
and there is no need for a separate term for the variance of $\delta$, as in Chu and Ghaharamani~\shortcite{chu2005preference}.
We assume a Gamma distribution $1/s \sim \mathcal{G}(a_0, b_0)$ with shape $a_0$ and scale $b_0$,
as this is a conjugate prior.

Given $N$ arguments and $P$ labeled preference pairs, $\mathbf y=\{y_1,...,y_P\}$,
we can make predictions by finding the posterior distribution over the convincingness values, 
$\mathbf f = \{f(\mathbf {x}_1),...,f(\mathbf {x}_N)\}$, given by:
\begin{flalign}
& p\left(\mathbf f | \mathbf{y}, k_{\theta}, a_0, b_0 \right) 
\propto p(\mathbf y | \mathbf f) p(\mathbf f | k_{\theta}, a_0, b_0) & \nonumber \\
& \! =  \frac{1}{Z} \! \int  \prod_{k=1}^P \Phi\!\left( z_k \right) 
\mathcal{N}(\mathbf f; \mathbf 0, \mathbf K_{\theta}s) \mathcal{G}(s; a_0, b_0) \mathrm{d}s, \!\!\!\! &
\label{eq:post}
\end{flalign}
where $Z = p\left(\mathbf{y} | k_{\theta}, a_0, b_0 \right)$.
Unfortunately, neither $Z$ nor the integral over $s$ 
can be computed analytically, so we must turn to approximations.

Chu and Ghahramani~\shortcite{chu2005preference}
used a Laplace approximation for GPPL, which finds a maximum a-posteriori (MAP) solution
that has been shown to perform poorly in many cases
~\cite{nickisch2008approximations}. 
More accurate estimates of the posterior could be obtained using Markov chain Monte Carlo sampling (MCMC),
but this is very computationally expensive ~\cite{nickisch2008approximations}. 
Instead, we use a faster \emph{variational} method that maintains the benefits of the Bayesian approach
~\cite{reece2011determining,steinberg2014extended} and adapt this method 
to the preference likelihood given by Equation \ref{eq:plphi}.

To apply the variational approach, we define an approximation $q(\mathbf f)$ to Equation \ref{eq:post}. 
First, we approximate the preference likelihood with a Gaussian, $\prod_{k=1}^P \Phi\left( z_k \right) \approx \mathcal{N}(\mathbf y; \mathbf G\hat{\mathbf f}, \mathbf Q)$. This allows us to avoid the intractable integral in $Z$ and obtain another Gaussian, $q(\mathbf f) = \mathcal{N}(\mathbf f; \hat{\mathbf f}, \mathbf C)$. 
The parameters $\hat{\mathbf f}$ and $\mathbf C$ 
depend on the approximate preference likelihood 
and an approximate distribution over $s$: $q(s) = \mathcal{G}(s; a, b)$. 
The variational inference algorithm begins by initializing the parameters $\mathbf G$, $ \hat{\mathbf f}$, $\mathbf C$, $a$ and $b$ at random. Then, the  algorithm proceeds iteratively updating each parameter in turn, given the current values for the other parameters. 
This optimization procedure minimizes the Kullback-Leibler (KL) divergence of $p(\mathbf f |\mathbf y, k_{\theta}, a_0, b_0)$ from $q(\mathbf f)$, causing $q(\mathbf f)$ to converge to an approximate posterior. 

The update equations for the mean $\hat{\mathbf f}$ and covariance $\mathbf C$ require inverting the covariance matrix, $K_{\theta}$, at a computational cost of $\mathcal{O}(N^3)$, which is impractical with more than a few hundred data points. 
Furthermore, the updates also require $\mathcal{O}(NP)$ computations and
have $\mathcal{O}(N^2 + NP + P^2)$ memory complexity.
To resolve this, 
we apply a recently introduced technique, stochastic variational inference (SVI) 
\cite{hoffman2013stochastic,hensman_scalable_2015},
to scale to datasets containing at least tens of thousands of arguments and pairwise labels.

SVI makes two approximations: it assumes $M$ \emph{inducing points},
which act as a substitute for the observed arguments;
it uses only a random subset of the data containing $P_n$ pairs at each iteration. 
At each iteration, $t$, rather than update $\hat{\mathbf{f}}$ and $\bs C$ directly, 
we update the mean $\hat{\mathbf{f}}_m$ and covariance $\bs C_m$ for the inducing
points. The update for each parameter $\lambda \in \{\hat{\mathbf{f}}_m, \bs C_m\}$ takes the form of a weighted mean of the previous estimate and a new estimate computed from only a subset of observations:
\begin{flalign}
\lambda^{(t)} = (1 - \rho_t) \lambda ^ {(t-1)} + \rho_t \hat{\lambda}_t P/P_n,
\end{flalign}
where $\rho=t^{-u}$ is the step size, $u$ is a forgetting rate, 
and $\hat{\lambda}_t$ is the new estimate computed from $P_n$ out of $P$ observations.
The values of $\hat{\mathbf{f}}$ and $\bs C$ can be estimated from 
the inducing point distribution.
By choosing $M <\!\!< N$ and $P_n <\!\!< P$, we limit the computational
complexity of each SVI iteration to $\mathcal{O}(M^3 + MP_n)$ and the 
memory complexity $\mathcal{O}(M^2 + MP_n + P_n^2)$.
To choose representative inducing points, 
we use K-means++~\cite{arthur2007k} with $K=M$ to rapidly cluster the feature vectors, 
then take the cluster centers as inducing points.
Compared to standard K-means, K-means++ introduces a new method for choosing the initial cluster seeds that
reduces the number of poor-quality clusterings.

A further benefit of GPs is that they enable \emph{automatic relevance determination (ARD)}
to identify informative features, which works as follows.
The prior covariance of $f$ is defined by a kernel function of the form 
$k_{\theta}(\mathbf x, \mathbf x') = \prod_{d=1}^D k_d(|x_d - x_d'| / l_d)$, 
where $k_d$ is a function of the distance between the values of feature $d$ 
for items $x$ and $x'$, and a length-scale hyper-parameter, $l_d$.
The length-scale controls the smoothness of the function across the feature space,
and can be optimized by choosing the value of $l_d$ that maximizes the approximate log marginal likelihood, $\mathcal{L} \approx \log p(\bs y)$. 
This process is known as \emph{maximum likelihood II (MLII)}~\cite{rasmussen_gaussian_2006}.
Features with larger length-scales after optimization are less relevant because their values
have less effect on $k_{\theta}(\mathbf x, \mathbf x') $.
To avoid the cost of optimizing the length-scales, we can alternatively set them using a median heuristic,
which has been shown to perform well in practice~\cite{gretton2012optimal}: 
$ \tilde{l}_{d} = \frac{1}{D} \mathrm{median}\left( \left\{ |x_{i,d} - x_{j,d}|, \right.\right.$
$ \left.\left. \forall i=1,...,N, \forall j=1,...,N\right\} \right) $.

\section{Experiments}\label{sec:expts}

\subsection{Datasets}
\begin{table*}[h]
\small
  \begin{tabularx}{\textwidth}{ p{2.0cm} | p{0.6cm} p{1.2cm} p{1.2cm} X }
  Dataset & Pairs & Arguments & Undecided & Dataset properties \\\hline\hline
  Toy Datasets & 4-13 & 4-5 & 0-9 & Synthetic pairwise labels
  \newline Arguments sampled at random from UKPConvArgStrict\\  
  \hline\emph{UKPConvArg-Strict} &
  11642 &
  1052 & 
  0 &
  Combine crowdsourced pairwise labels with MACE \newline
  Gold labels are $\ge 95\%$ most confident MACE labels \newline
  Discard arguments marked as equally convincing \newline
  Discard conflicting preferences \\
  \hline\emph{UKPConvArg-Rank} &
  16081 &
  1052 &
  3289 &
  Combine crowdsourced pairwise labels with MACE \newline
  Gold labels are $\ge 95\%$ most confident MACE labels \newline
  PageRank run on each topic to produce gold rankings \\  
  \hline\emph{UKPConvArg-CrowdSample} &
  16927 & 
  1052 &
  3698 &
  One original crowdsourced label per pair\newline
  PageRank run on each topic to produce gold rankings \newline
  Labels for evaluation from UKPConvArgStrict/UKPConvArgRank
  \end{tabularx}
  \caption{\label{tab:expt_data} Summary of datasets, showing the different steps used to produce each Internet argument dataset.}
\end{table*}
We first use toy datasets to illustrate the behavior of several different methods (described below).
Then, 
we analyze the scalability and performance of our approach on datasets provided by Habernal and Gurevych~\shortcite{habernal2016argument},
which contain pairwise labels for arguments taken from online discussion forums.
The labels can have a value of $0$, meaning the annotator found the second argument in the pair more convincing,
$1$ if the annotator was undecided, or $2$ if the first argument was more convincing.
To test different scenarios, different pre-processing steps were used to produce the
three \emph{UKPConvArg*} datasets shown in Table \ref{tab:expt_data}.
\emph{UKPConvArgStrict} and \emph{UKPConvArgRank} were cleaned to remove disagreements between annotators, hence can be considered to be \emph{noise-free}.
 \emph{UKPConvArgCrowdSample} is used to evaluate performance with noisy crowdsourced data 
including conflicts and undecided labels, and to test the suitability of our method for active learning
to address the cold-start problem in domains with no labeled data.
For these datasets, we perform 32-fold cross validation, where
each fold corresponds to one of two stances for one of 16 controversial topics.
 
\subsection{Method Comparison}

\begin{figure*}
\centering
\subfloat[no cycle]{
  \includegraphics[width=.35\columnwidth, clip=True, trim=30 50 20 48]{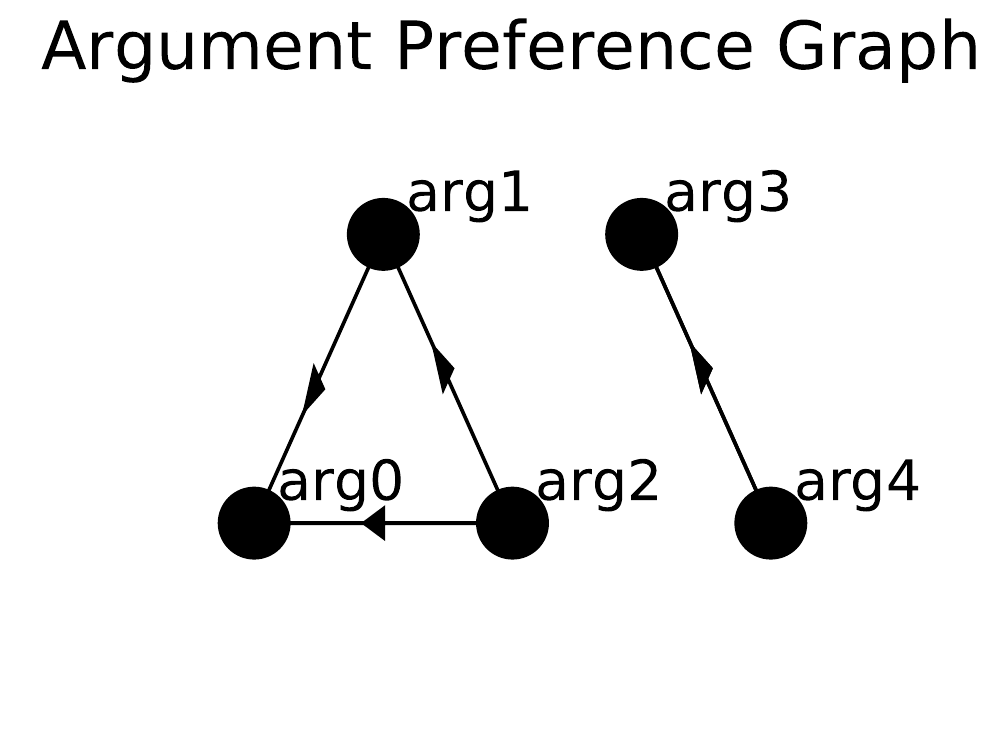}
}
\subfloat[single cycle]{
  \includegraphics[width=.35\columnwidth, clip=True, trim=30 50 20 48]{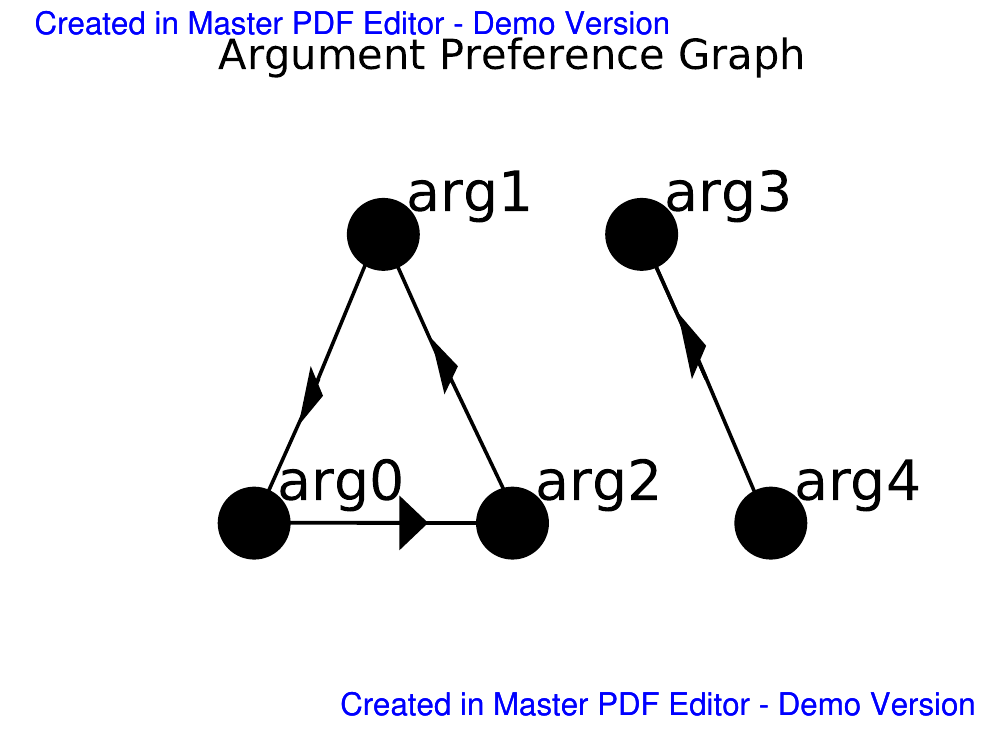}
}
\subfloat[double cycle]{
  \includegraphics[width=.37\columnwidth, clip=True, trim=20 50 20 48]{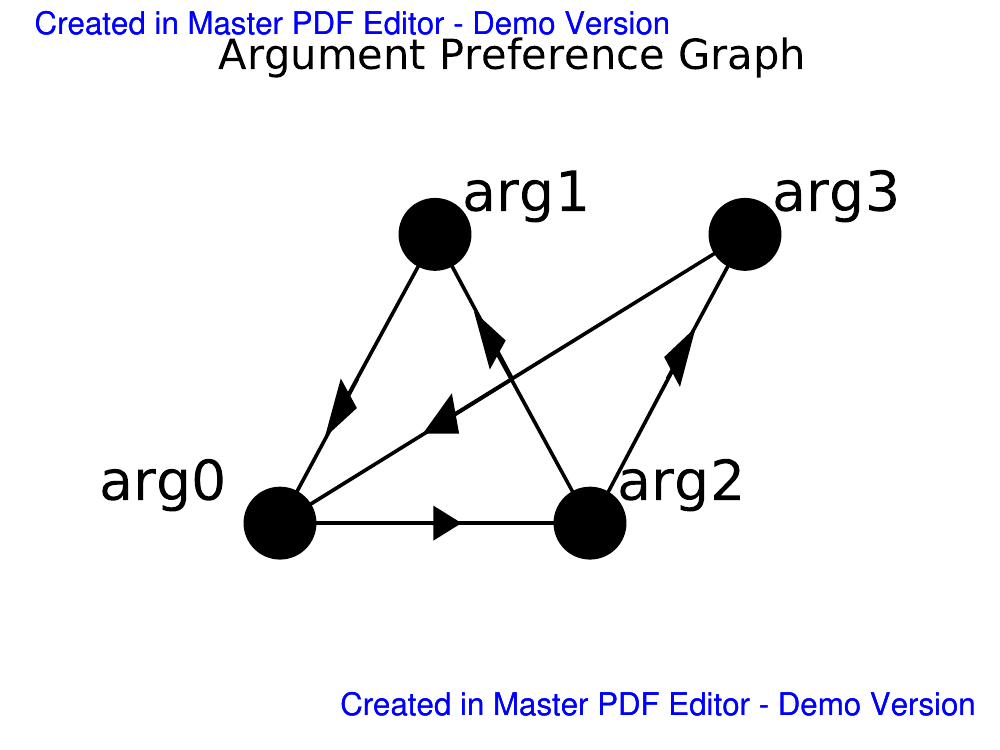}
}
\subfloat[no cycle + 9 undecided prefs.]{
  \includegraphics[width=.5\columnwidth, clip=True, trim=-30 50 -30 48]{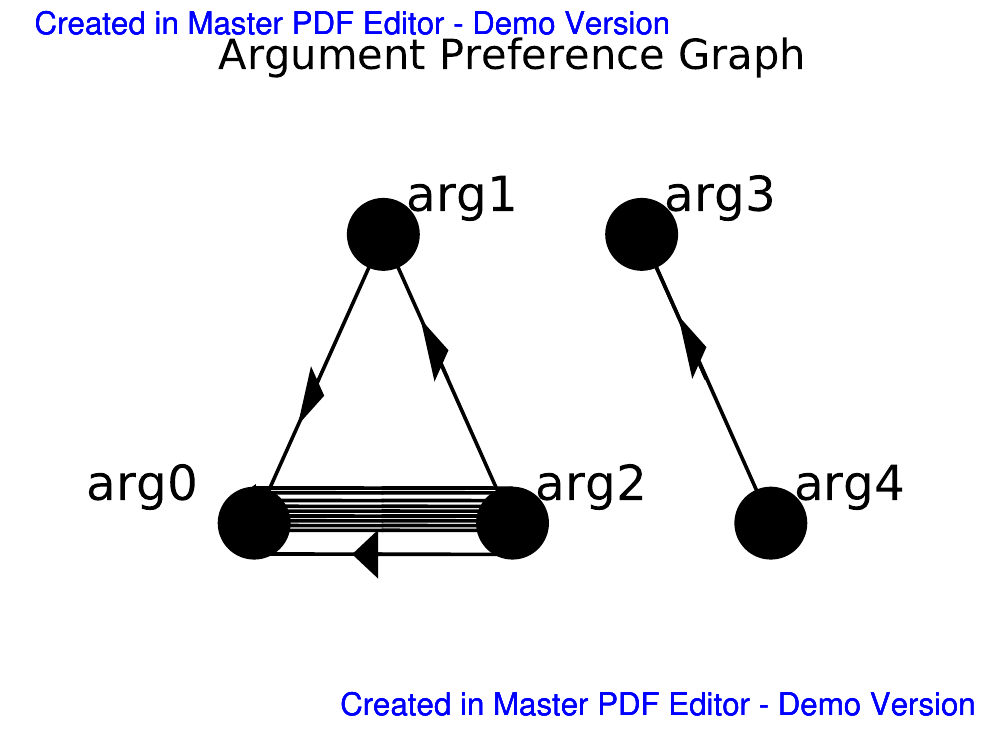}
}
\caption{Argument preference graphs for each scenario. Arrows point to the preferred argument.}
\label{fig:arg_graph}
\end{figure*}
\begin{figure*}
\centering
\subfloat[no cycle]{
  \includegraphics[width=.40\columnwidth, clip=True, trim=20 47 10 25]{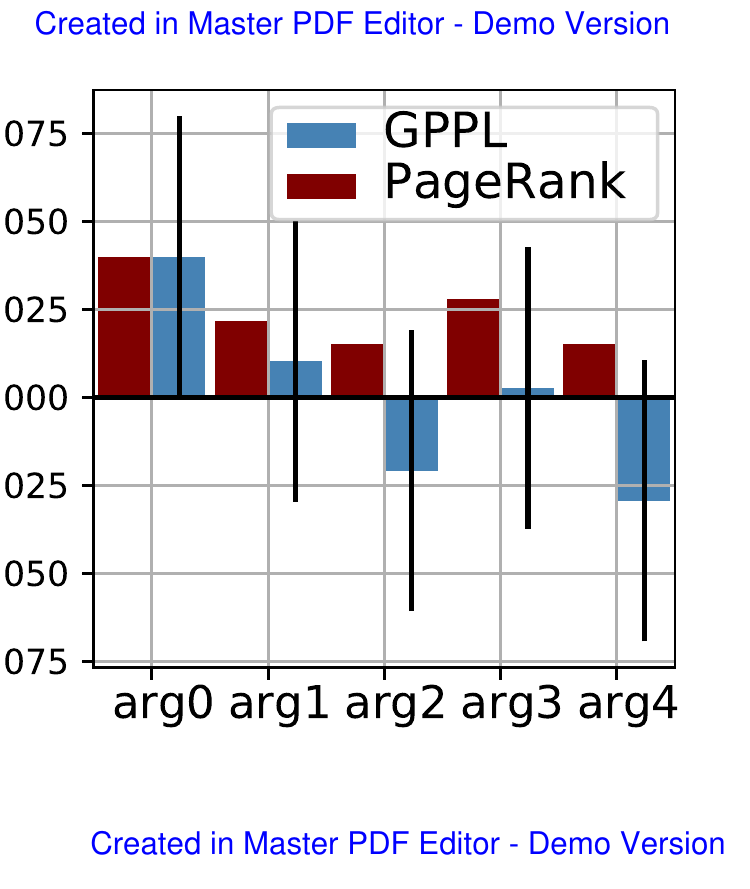}
}
\subfloat[single cycle]{
  \includegraphics[width=.40\columnwidth, clip=True, trim=20 47 10 22]{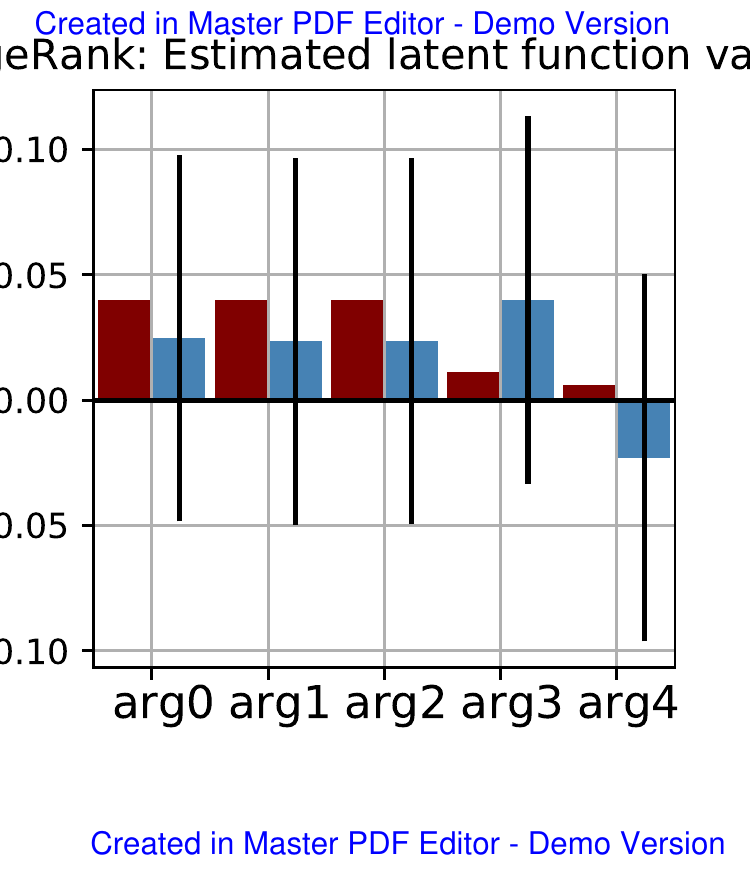}
}
\subfloat[double cycle]{
  \includegraphics[width=.40\columnwidth, clip=True, trim=20 47 10 22]{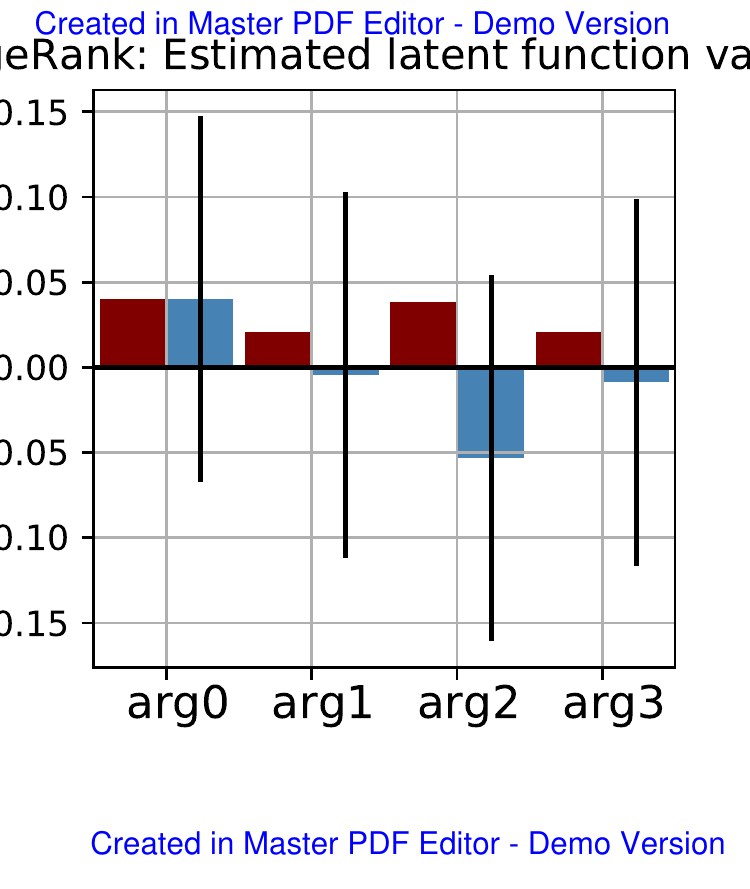}
}
\subfloat[9 undecided]{
  \includegraphics[width=.40\columnwidth, clip=True, trim=20 47 10 22]{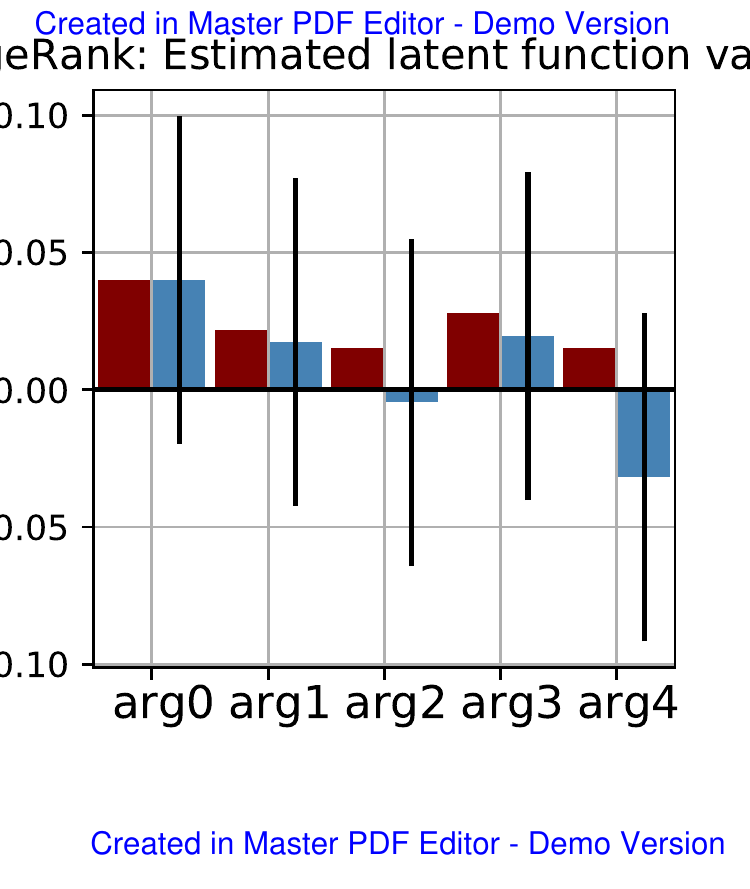}
  \label{fig:ugraph}
}
\caption{Mean scores over 25 repeats. Bars for GPPL show standard deviation of convincingness function posterior.}
\label{fig:scores}
\end{figure*}
Our two tasks are \emph{ranking} arguments by convincingness and  
\emph{classification} of pairwise labels to predict which argument is more convincing. 
For both tasks, our proposed GPPL method is trained using the pairwise labels for the training folds.
We rank arguments by their expected convincingness, $\mathbb{E}[f(\mathbf{x}_i)]\approx \hat{f}(\mathbf {x_i})$ for each argument $i$ with feature vector $\mathbf{x}_i$, under the approximate posterior $q(\mathbf f)$
output by our SVI algorithm.
We obtain classification probabilities using Equation \ref{eq:plphi} but 
accommodate the posterior covariance, $\mathbf C$, of $\mathbf f$, by replacing $z$ with $\hat{z} = (\hat{f}(\mathbf x_i) - \hat{f}(\mathbf x_j)) / \sqrt{2 + C_{ii} + C_{jj} - C_{ij} - C_{ji}}$.
We tested the sensitivity of GPPL to the choice of seed values for K-means++ by training the model on the same $31$ folds of UKPConvArgStrict $20$ times, each with a different random seed, then testing on the remaining fold.
The resulting accuracy had a standard deviation of $0.03$. 
In the following experiments, all methods were initialized and trained once for each fold of each experiment.

We compare GPPL to an SVM with radial basis function kernel, 
and a bi-directional long short-term memory network (BiLSTM),
with $64$ output nodes in the core LSTM layer. 
The SVM and BiLSTM were tested by Habernal and Gurevych~\shortcite{habernal2016argument} and are available in our software repository.
To apply SVM and BiLSTM to the classification task, we concatenate the feature vectors of each pair of arguments and train on the pairwise labels.
For ranking, PageRank is first applied to arguments in the training folds to obtain scores from the pairwise labels,
which are then used to train the SVM and BiLSTM regression models.

As a Bayesian alternative to GPPL, 
we test a Gaussian process classifier (\emph{GPC}) for the classification task 
by concatenating the feature vectors of arguments in the same way as the SVM classifier.
We also evaluate a non-Bayesian approach that infers function values using the 
same pairwise preference likelihood (\emph{PL}) as GPPL
(Equation \ref{eq:plphi}), 
but uses them to train an SVM regression model instead of a GP. 
We refer to this method as \emph{PL+SVR}.

We use two sets of input features. The \emph{ling} feature set contains $32,010$ linguistic features,  
including unigrams, bigrams, parts-of-speech (POS) n-grams, production rules,
ratios and counts of word length, punctuation and verb forms,
dependency tree depth, named entity type counts,
readability measures, sentiment scores, and spell-checking.
The \emph{GloVe} features are word embeddings with 300 dimensions. Both feature sets were
developed by Habernal and Gurevych~\shortcite{habernal2016argument}.
We also evaluate a combination of both feature sets, \emph{ling + GloVe}.
To create a single embedding vector per argument as input for GPPL,
we take the mean of individual word embeddings for tokens in the argument.
We also tested skip-thoughts~\cite{kiros2015skip} and Siamese-CBOW~\cite{kenter2016siamesecbow} 
with GPPL on UKPConvArgStrict and UKPConvArgRank, both with MLII optimization and the median heuristic,
 both alone and combined with \emph{ling}. 
However, we found that mean GloVe embeddings produced substantially better performance in all tests.
To input the argument-level \emph{ling} features to BiLSTM, we extend the network by adding a dense layer with $64$ nodes. 

We set the GPPL hyper-parameters $a_0=2$ and $b_0=200$ by comparing
training set performance on UKPConvArgStrict and UKPConvArgRank against $a_0=2$, $b_0=20000$ and $a_0=2$, $b_0=2$.
The chosen prior is very weakly informative, favoring a moderate level of noise in the pairwise labels.
For the kernel function, $k_d$, we used the 
Mat\'ern $\frac{3}{2}$ function as it has been shown to outperform 
other commonly-used kernels, such as RBF, across a wide range of tasks~\cite{rasmussen_gaussian_2006}.
We defer evaluating other kernel functions to future work.
To set length-scales, $l_d$, we compare the median heuristic (labeled ``medi.")
with MLII optimization using an L-BFGS optimizer (``opt."). Experiment 2 shows how
the number of inducing points, $M$, can be set to trade off speed and accuracy. 
Following those results, we set $M=500$ for Experiments 3, 4 and 5 and $M=N$ for the toy dataset in Experiment 1.

\subsection{Experiment 1: Toy Data}

\begin{figure}
\centering
\captionsetup[subfloat]{labelformat=empty}
\subfloat[\;no cycle]{
  \includegraphics[width=.22\columnwidth, clip=True, trim=58 20 47 24]{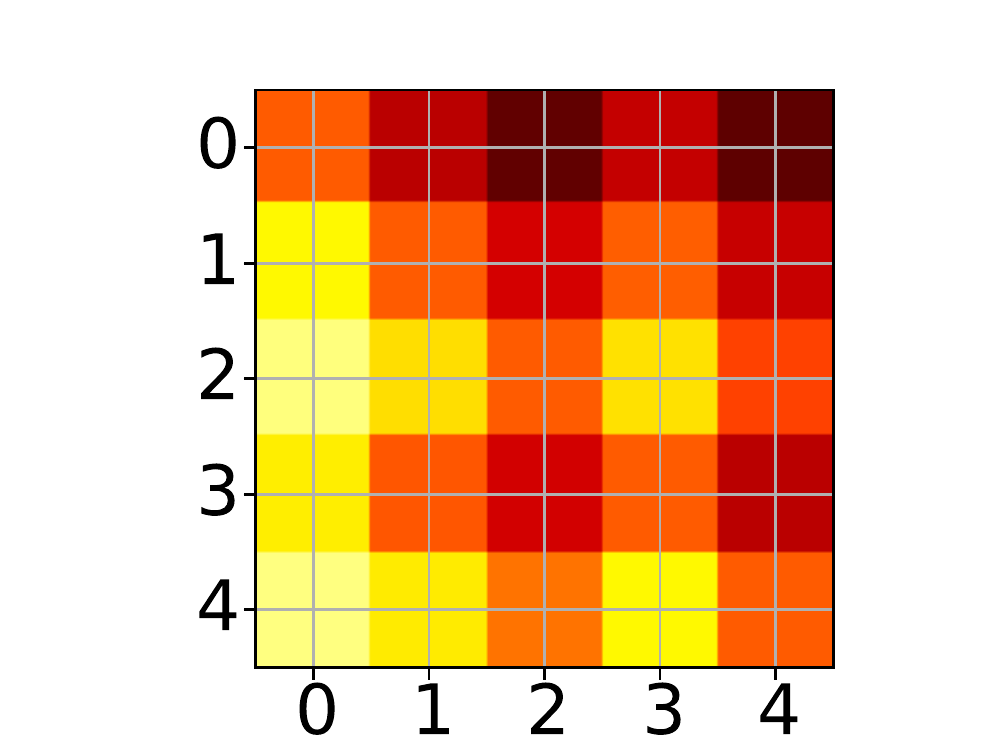}
}
\subfloat[single cycle]{
  \includegraphics[width=.205\columnwidth, clip=True, trim=70 20 48 24]{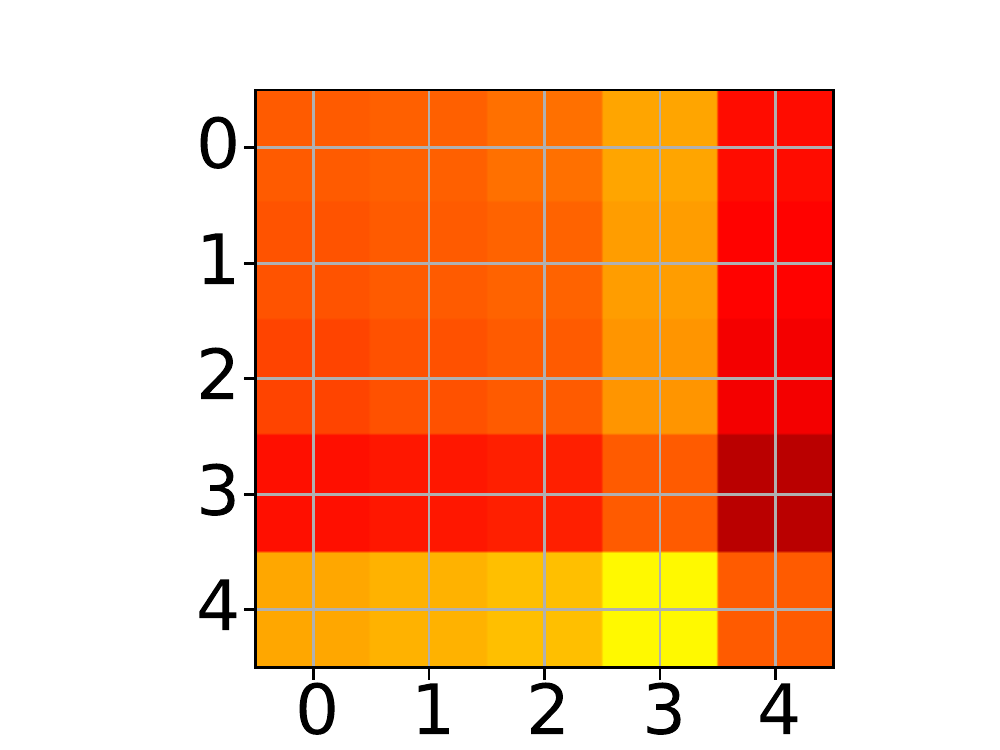}
}
\subfloat[double cycle]{
  \includegraphics[width=.205\columnwidth, clip=True, trim=70 20 48 24]{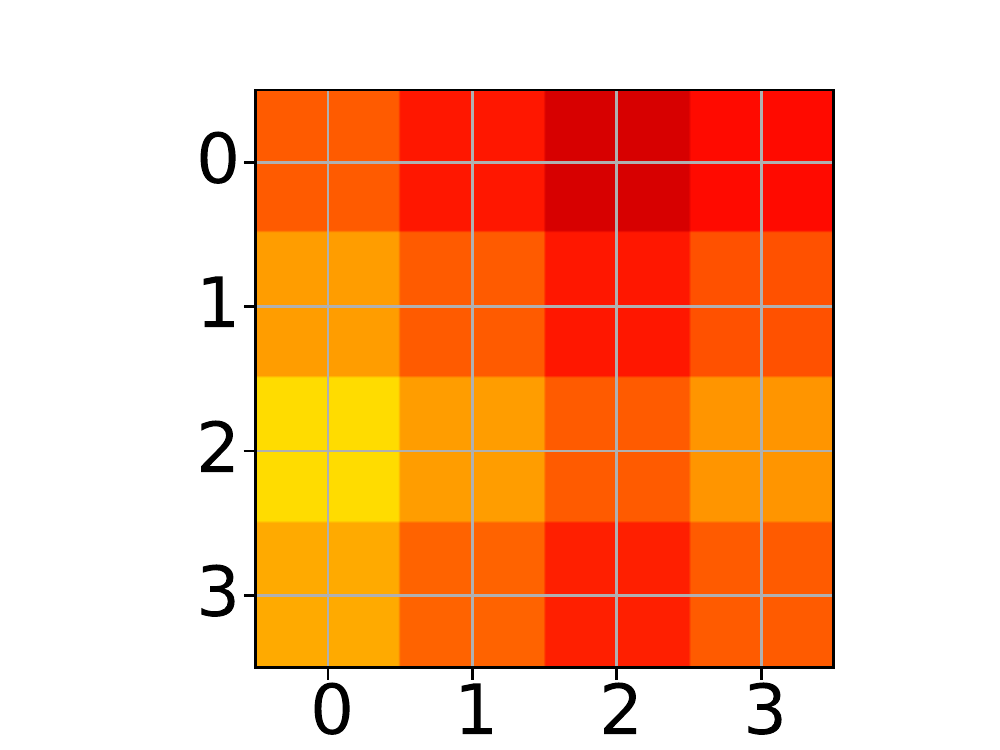}
}
\subfloat[9 undecided\;\;\;\;\;]{
  \includegraphics[width=.278\columnwidth, clip=True, trim=55 20 4 20]{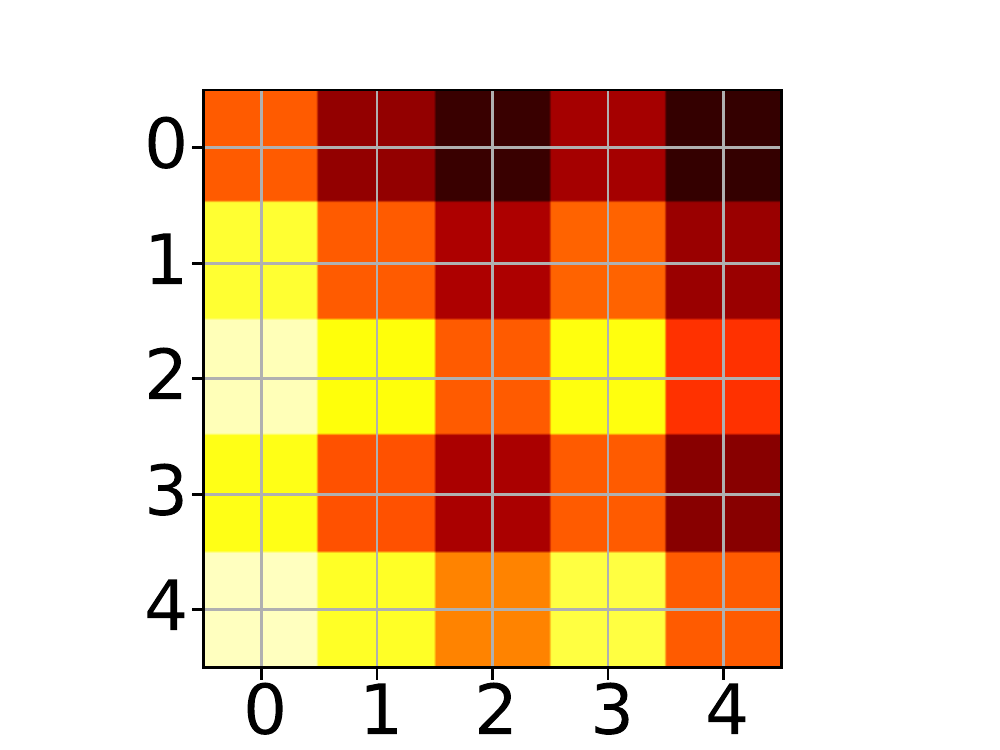}
}\\[-12pt]
\subfloat[]{
  \includegraphics[width=.22\columnwidth, clip=True, trim=58 5 47 24]{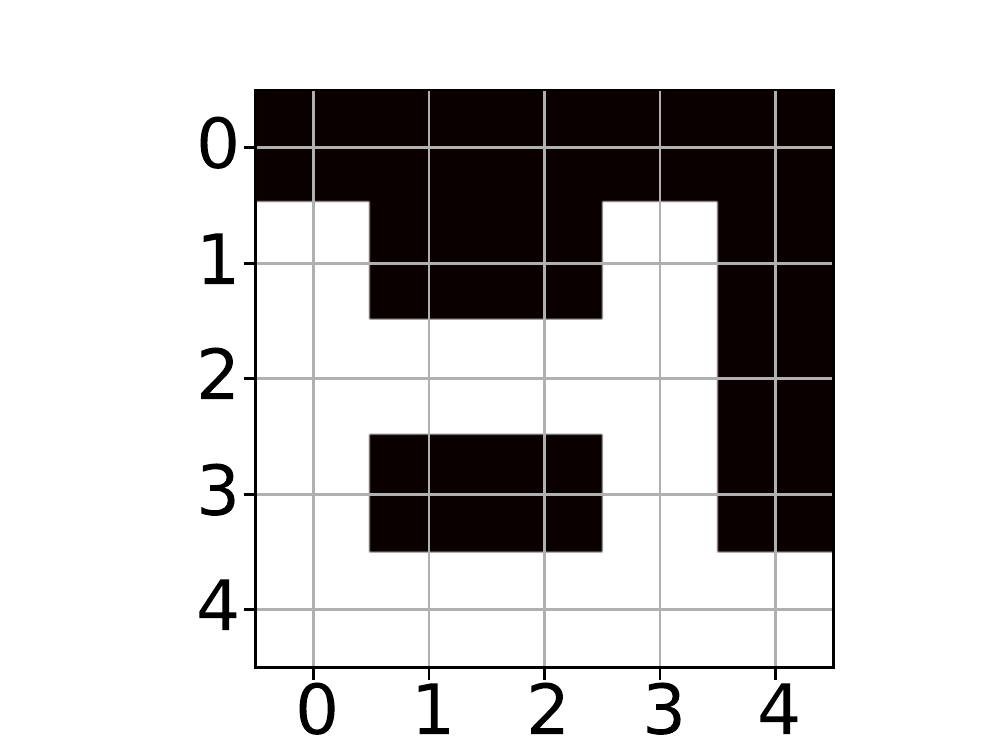} 
}
\subfloat[]{
  \includegraphics[width=.205\columnwidth, clip=True, trim=70 5 48 24]{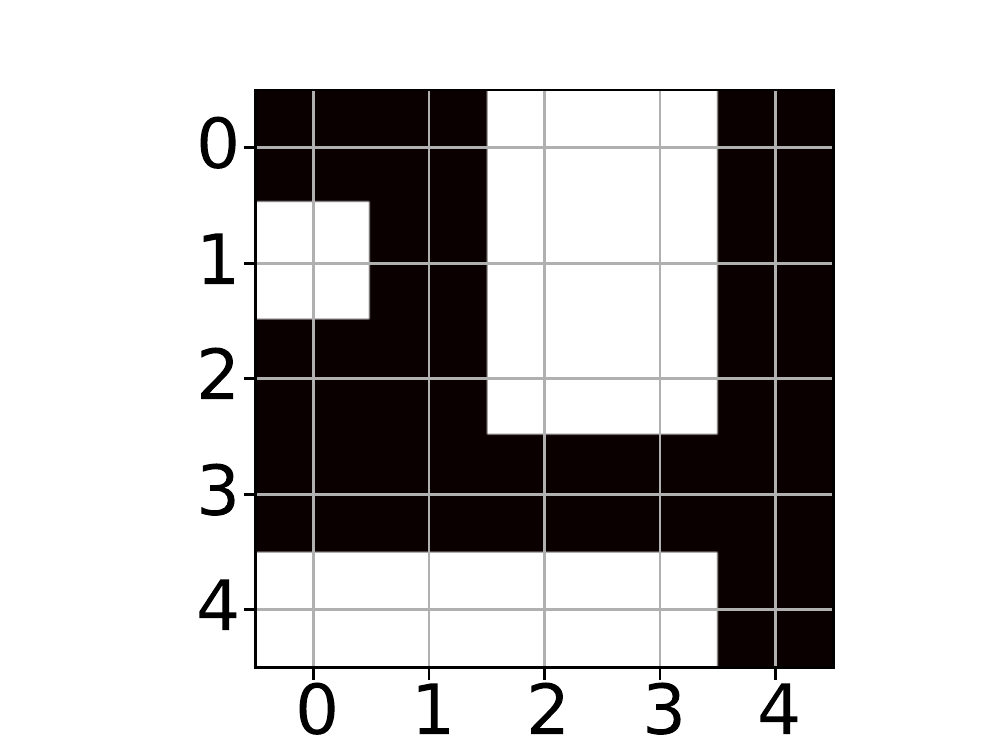} 
}
\subfloat[]{
  \includegraphics[width=.205\columnwidth, clip=True, trim=70 5 48 24]{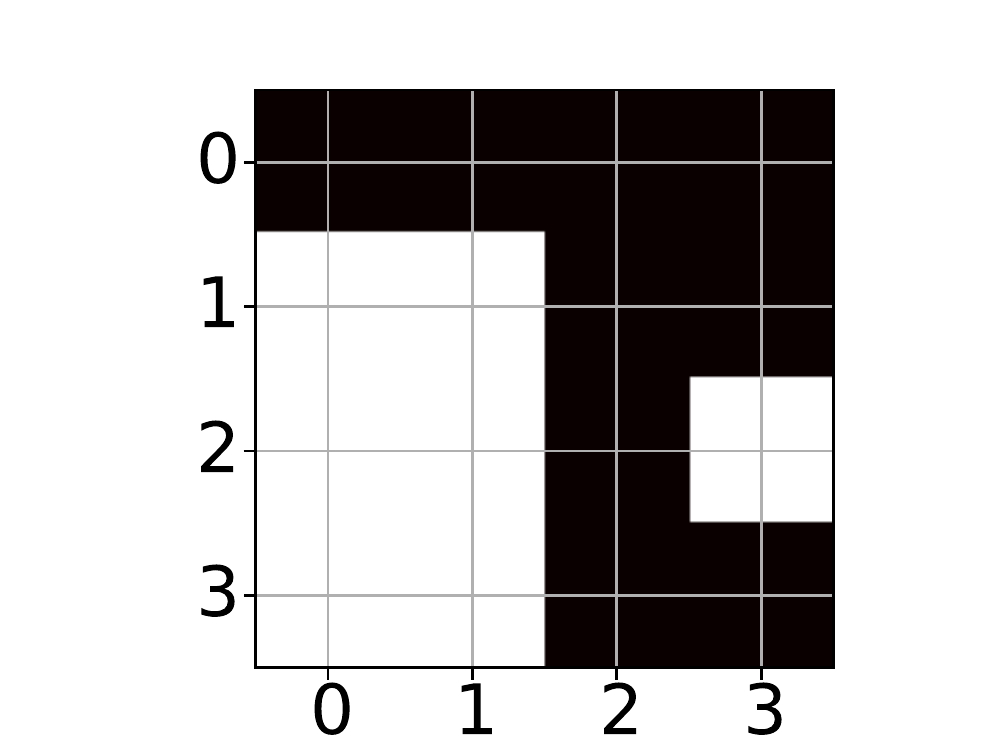} 
}
\subfloat[]{
  \includegraphics[width=.278\columnwidth, clip=True, trim=55 5 4 20]{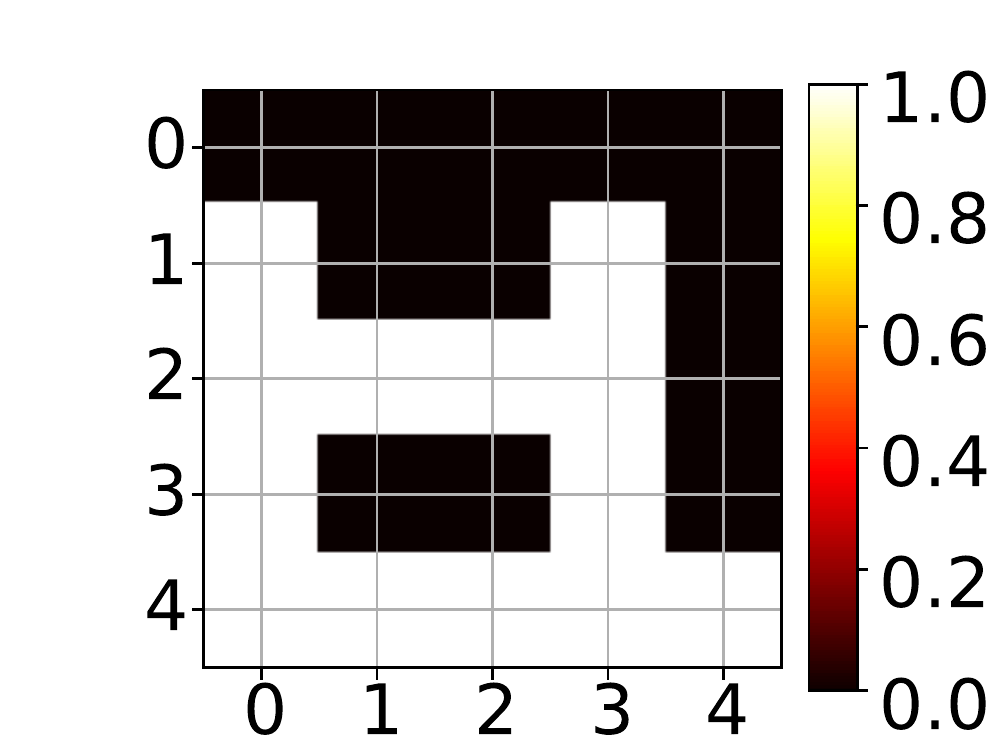} 
}\\[-12pt]
\caption{Mean GPPL (top row) and SVM (bottom row) predictions over 25 repeats. Probability that the argument 
on the horizontal axis $\succ$ the argument on the vertical axis.}
\label{fig:cycle_demo_classification}
\end{figure}

To illustrate some key differences between GPPL, SVM and PageRank,
we simulate four scenarios, each of which contains arguments labeled \emph{arg0} to \emph{arg4}.  
In each scenario, we generate a set of pairwise preference labels according to the 
graphs shown in Figure \ref{fig:arg_graph}.
Each scenario is repeated 25 times: in each repeat, we select arguments at random from one fold of UKPConvArgStrict
then associate the mean GloVe embeddings for these arguments with the labels arg0 to arg4. 
We train GPPL, PageRank and the SVM classifier on the preference pairs shown in each graph and
predict ranks and pairwise labels for arguments arg0 to arg4.

\begin{figure*}[h]
\centering
\subfloat[Varying no. arguments in training set, GloVe features]{
\includegraphics[width=0.62\columnwidth, clip=True, trim=12 18 0 13.5]{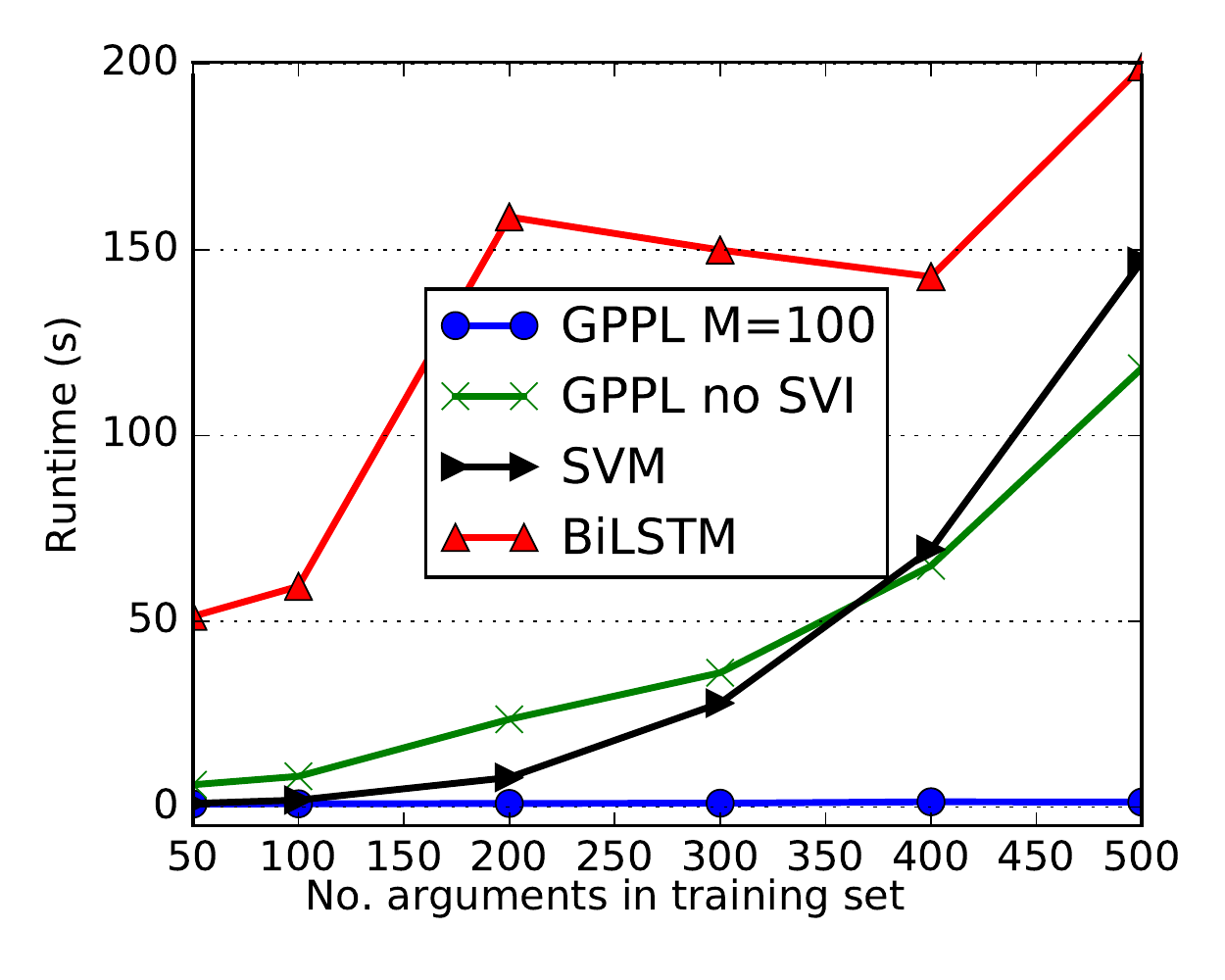}
\label{fig:scale_N}
}
\hspace{0.1cm}
\subfloat[Varying no. ling+GloVe features, GPPL, medi., M=500]{
\includegraphics[width=0.55\columnwidth, clip=True, trim=20 47 0 20]{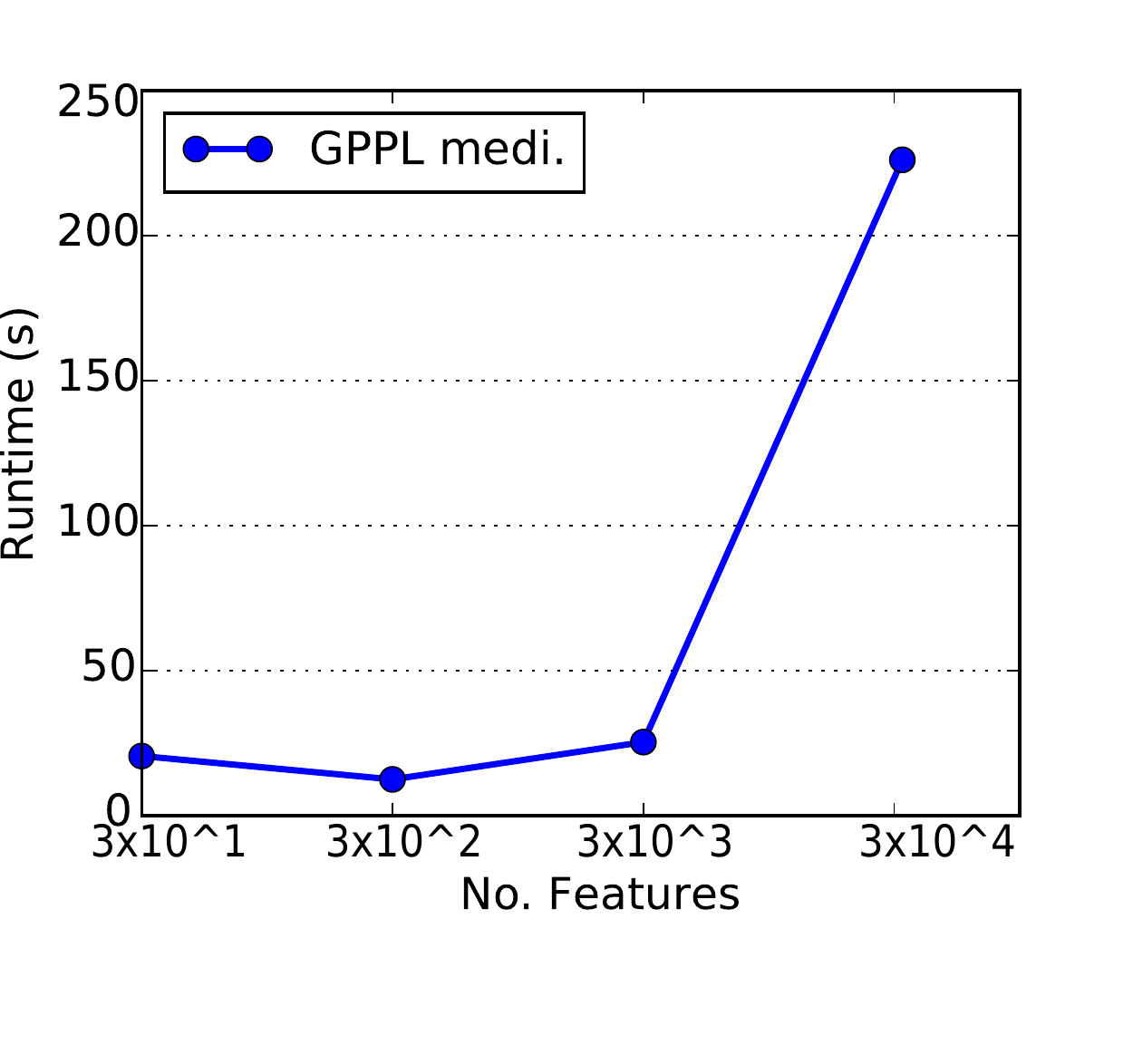}
\label{fig:scale_dims}
}
\hspace{0.1cm}
\subfloat[Varying no. ling+GloVe features, long-running methods]{
\includegraphics[width=0.58\columnwidth, clip=True, trim=0 58 10 20]{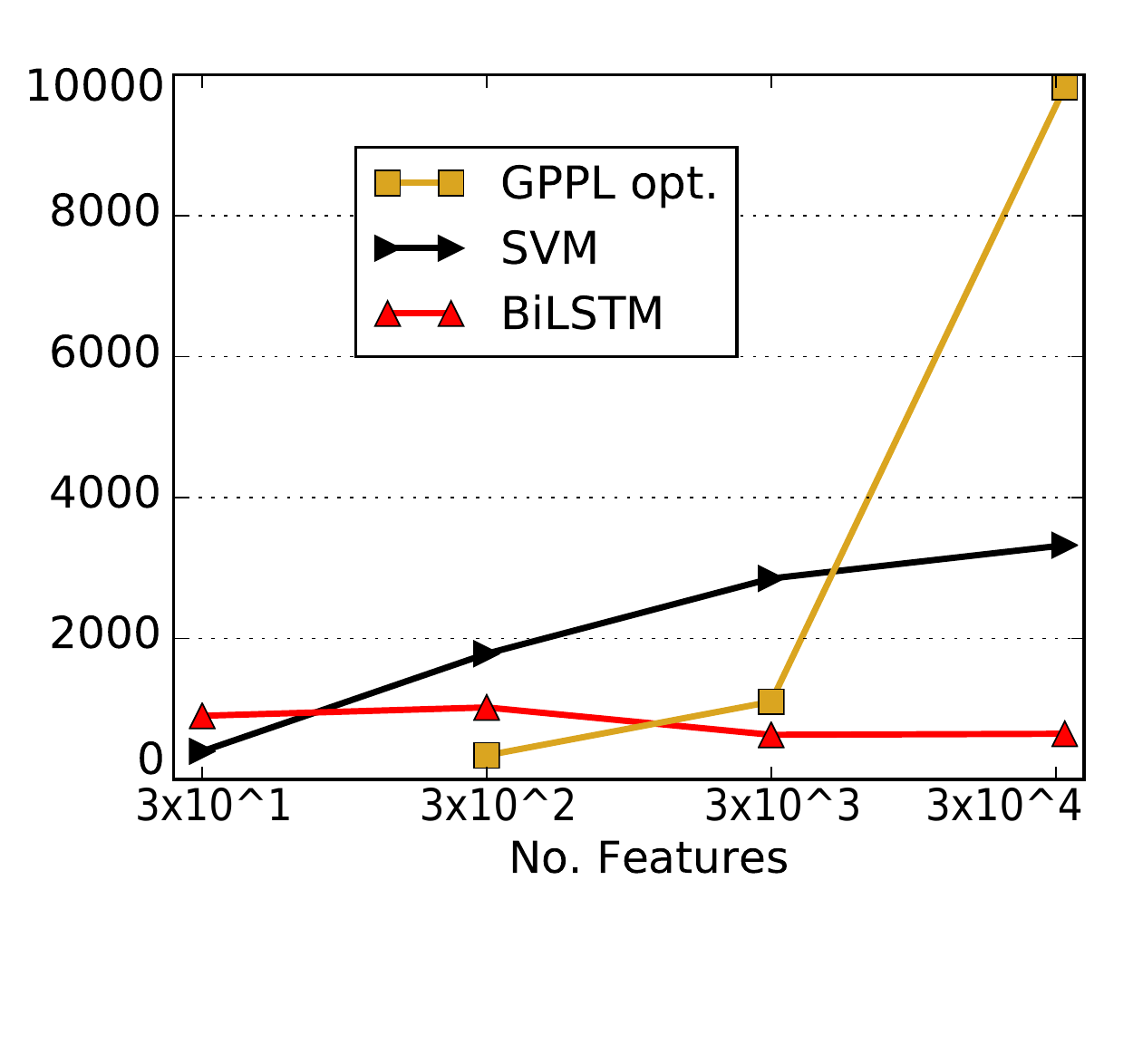}
\label{fig:scale_dims_others}
}
\caption{Runtimes for training+prediction on UKPConvArgStrict with different subsamples of data. Means over 32 runs. Note logarithmic x-axis for (b) and (c). }
\end{figure*} 
\begin{figure}[t]
\subfloat[33210 ling+GloVe features]{
\hspace{-1.5mm}
\includegraphics[width=0.49\columnwidth, clip=True, trim=4 40 13 12]{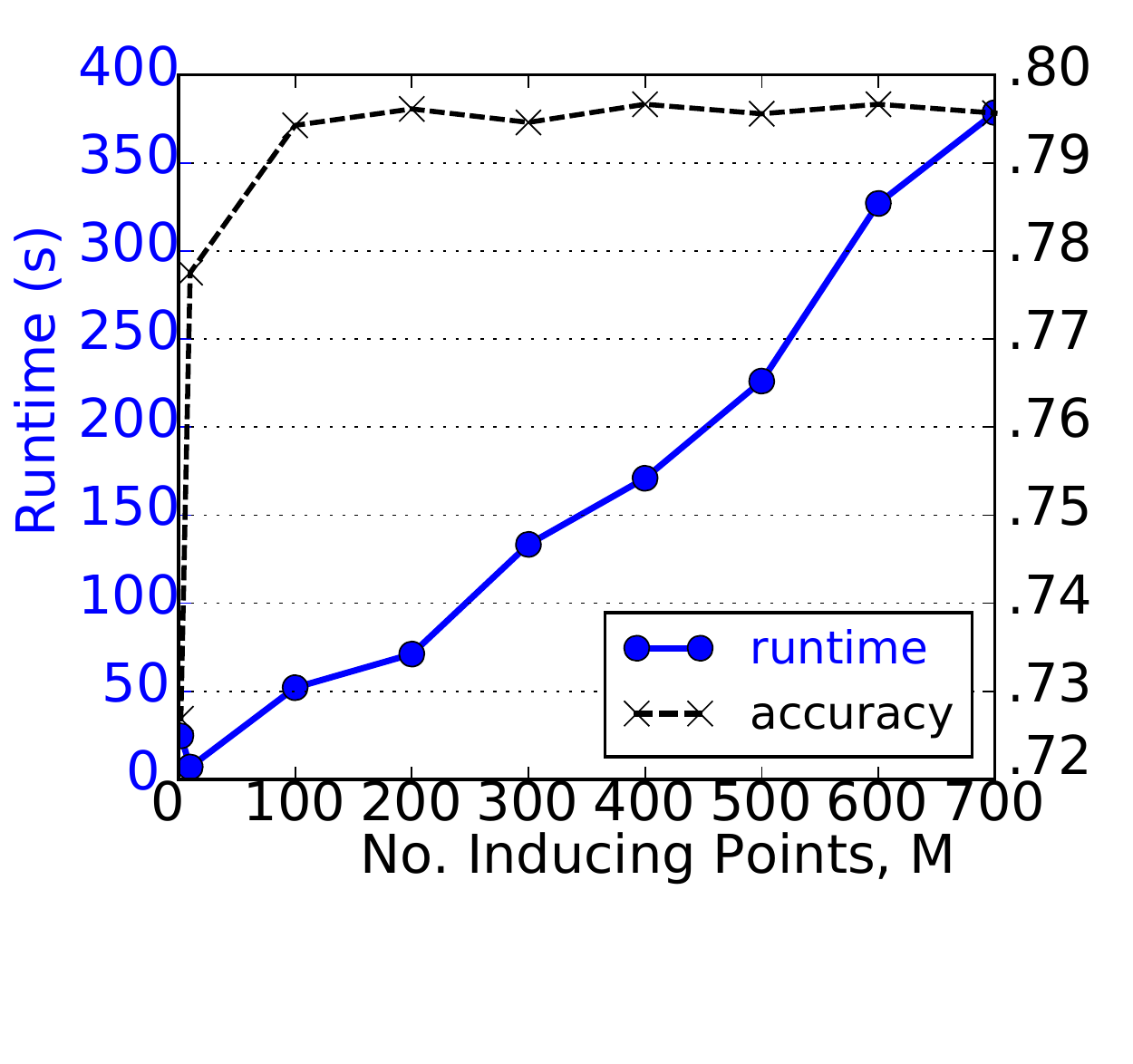}
\label{fig:scale_M_b}}
\subfloat[300 GloVe features]{
\hspace{-1.5mm}
\includegraphics[width=0.48\columnwidth, clip=True, trim=8 40 17 12]{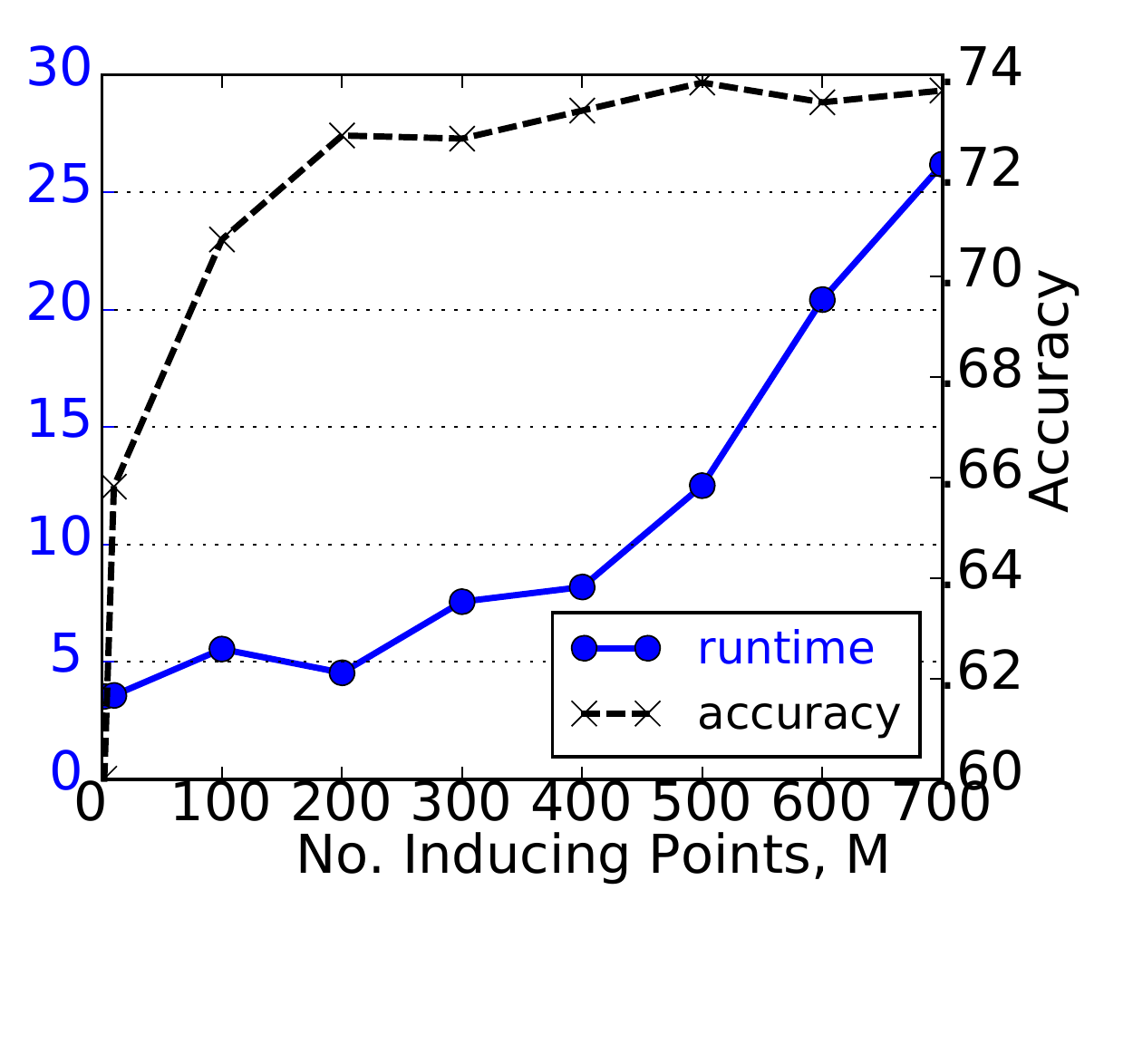}
\label{fig:scale_M_a}}
\caption{Effect of varying $M$ on accuracy and runtime (training+prediction) of GPPL for UKPConvArgStrict.  Means over 32 runs.}
\label{fig:scale_M}
\end{figure}
In the  ``no cycle" scenario, 
arg0 is preferred to both arg1 and arg2, which is reflected in the scores predicted by PageRank and GPPL in Figure \ref{fig:scores}. However, arg3 and arg4 are not connected to the rest of the graph, and PageRank and GPPL score them differently. 
Figure \ref{fig:cycle_demo_classification} shows how GPPL provides less confident classifications for pairs that were not yet observed, e.g. arg2 $\succ$ arg4, in contrast with the discrete classifications of the SVM.

The next scenario shows a ``single cycle" in the preference graph.
Both PageRank and GPPL produce equal values for the arguments in the cycle (arg0, arg1, arg2). PageRank assigns lower scores to both arg3 and arg4 than the arguments in the cycle, 
while GPPL more intuitively gives a higher score to arg3, which was preferred to arg4. 
SVM predicts that arg0 and arg1 are preferred over arg3, 
although arg0 and arg1 are in a cycle so there is no reason to prefer them. 
GPPL, in contrast,  weakly predicts that arg3 is preferred.

The ``double cycle" scenario contains two paths from arg2 to arg0, via arg1 or arg3, and one conflicting
preference arg2 $\succ$ arg0. 
GPPL scores the arguments as if the single conflicting preference, arg2 $\succ$ arg0, 
is less important than the two parallel paths from arg2 to arg0. 
In contrast, PageRank gives high scores to both arg0 and arg2.
The classifications by GPPL and SVM are similar, but GPPL produces more uncertain 
predictions than in the first scenario due to the conflict.

Finally,  Figure \ref{fig:ugraph} shows the addition of $9$ undecided labels to the ``no cycle" scenario, indicated by 
undirected edges in Figure \ref{fig:arg_graph}, to simulate multiple annotators viewing the pair without being able to choose the most convincing argument.
The SVM and PageRank are unaffected as they cannot be trained using the undecided labels.
However, the GPPL classifications are less confident and the difference in GPPL scores between arg0 and the other arguments decreases, since GPPL gives the edge from arg2 to arg0 less weight.

In conclusion, GPPL appears to resolve conflicts in the preference graphs
more intuitively than PageRank, which was designed to rank web pages by 
importance rather than preference. 
In contrast to SVM, GPPL is able to account for cycles and undecided labels to soften its predictions.

\subsection{Experiment 2: Scalability}

We analyze empirically the scalability of the proposed SVI method for GPPL using the UKPConvArgStrict dataset.
Figure \ref{fig:scale_M} shows the effect of varying the number of inducing points, $M$, on the overall runtime and accuracy of the method. The accuracy increases quickly with $M$, and flattens out, suggesting there is little benefit to increasing  $M$ further on this dataset. 
The runtimes increase with $M$,  and are much longer with $32,310$ features than with 300 features.
The difference is due to the cost of computing the kernel, which is linear in $M$,
With only $300$ features, the Figure \ref{fig:scale_M_a} runtime appears polynomial, reflecting the 
$\mathcal{O}(M^3)$ term in the inference procedure. 

\begin{table*}
\small
  \begin{tabularx}{\textwidth}{ | l | X | X | X |  X |  X |  X |  X | X | X | X |}
  \hline
       &\multicolumn{2}{c|}{SVM}&\multicolumn{2}{c|}{BiLSTM}&\multicolumn{3}{c|}{GPPL median heuristic}&GPPL opt. & GPC & PL+ SVR\\\hline
       &ling &ling +GloVe &GloVe &ling +GloVe &ling &GloVe &\multicolumn{4}{c|}{ling +GloVe}\\\hline
\multicolumn{11}{| l |}{UKPConvArgStrict (pairwise classification)} \\   \hline       
Accuracy  &.78 & .79 &.76 & .77 &.78 &.71  &.79  & .80 & \textbf{.81} & .78\\
ROC AUC   &.83 & .86 &.84 & .86 &.85 &.77  &.87  & .87 & \textbf{.89} & .85\\
CEE   &.52 & .47 &.64 & .57 &.51 &1.12  &.47  & .51 & \textbf{.43} & .51 \\
\hline \multicolumn{11}{| l |}{UKPConvArgRank (ranking)} \\   \hline
Pearson's r      &.36 & .37 &.32 & .36 &.38 &.33  & \textbf{.45} &  .44 & - & .39 \\
Spearman's $\rho$&.47 & .48 &.37 & .43 &.62 &.44  &.65&  \textbf{.67} & - & .63\\
Kendall's $\tau$ &.34 & .34 &.27 & .31 &.47 &.31  &.49   &  \textbf{.50} & - & .47\\
\hline
  \end{tabularx}
  \caption{Performance comparison on UKPConvArgStrict and UKPConvArgRank datasets. }
  \label{tab:clean_results}
\end{table*}
We tested GPPL with both the SVI algorithm, with $M=100$ and $P_n=200$, and variational inference without inducing points or stochastic updates (labeled ``no SVI'') with different sizes of training dataset subsampled from UKPConvArgStrict. 
The results are shown in Figure \ref{fig:scale_N}. 
For GPPL with SVI, the runtime increases very little with dataset size, 
while the runtime with ``no SVI'' increases polynomially with training set size (both $N$ and $P$). 
At $N=100$, the number of inducing points is $M=N$ but the SVI algorithm is still faster due to the stochastic updates with $P_n=200 \ll P$ pairs.

Figure \ref{fig:scale_dims} shows the effect of the number of features, $D$, on runtimes.  
Runtimes for GPPL increase by a large amount with $D=32,310$,
 because the SVI method computes the kernel matrix, $K_{mm}$, with computational complexity $\mathcal{O}(D)$. 
 While $D$ is small, other costs dominate. 
We show runtimes using the MLII optimization procedure with GPPL in Figure \ref{fig:scale_dims_others}. 
Owing to the long computation times required, the procedure was limited to
a maximum of 25 iterations and did not terminate in fewer than 25 in any of the test runs. 
This creates a similar pattern to Figure \ref{fig:scale_dims} (approximately multiples of $50$).

We include runtimes for SVM and BiLSTM in Figures \ref{fig:scale_N} and
\ref{fig:scale_dims_others} to show their runtime patterns, but note that the runtimes reflect differences in implementations and system hardware.
Both SVM and GPPL were run on an Intel i7 quad-core desktop. For SVM we used LibSVM version 3.2, which could be sped up if probability estimates were not required.
BiLSTM was run with Theano 0.7\footnote{\url{http://deeplearning.net/software/theano/}} on an Nvidia Tesla P100 GPU. 
We can see in Figure \ref{fig:scale_dims_others} that the runtime for BiLSTM does
not appear to increase due to the number of features, while that of SVM increases sharply with $32,310$ features. 
In Figure \ref{fig:scale_N}, we observe the SVM runtimes increase polynomially with training set size. 

\subsection{Experiment 3: UKPConvArgStrict and UKPConvArgRank}

We compare classification performance on UKPConvArgStrict  
and ranking performance on UKPConvArgRank. 
The results in Table \ref{tab:clean_results} show that when using \emph{ling} features,
GPPL produces similar accuracy and improves the area under the ROC curve (AUC) by $.02$ and cross entropy error (CEE) by $.01$.
AUC quantifies how well the predicted probabilities separate the classes,
while CEE quantifies the usefulness of the probabilities output by each method.
Much larger improvements can be seen in the ranking metrics. 
When GPPL is run with \emph{GloVe}, it performs worse than
BiLSTM for classification but improves the ranking metrics. 
Using a combination of features improves all methods, suggesting that embeddings and linguistic features contain complementary information. This improvement is statistically significant ($p \ll .01$ using two-tailed Wilcoxon signed-rank test) for SVM with all metrics except accuracy, for BiLSTM with AUC only, and for GPPL medi. with Pearson correlation only.

Optimizing the length-scale using MLII improves classification accuracy by 1\% over the median heuristic,
and significantly improves accuracy ($p=.043$) and AUC ($p=.013$) 
over the previous state-of-the-art, SVM \emph{ling}.
However, the cost of these improvements is that each fold required around 2 hours to compute instead of 
approximately 10 minutes on the same machine (Intel i7 quad-core desktop) using the median heuristic. 
The differences in all ranking metrics between GPPL opt. and SVM \emph{ling + GloVe} 
are statistically significant, with $p=.029$ for Pearson's $r$ and $p\ll.01$ for both 
Spearman's $\rho$ and Kendall's $\tau$.

GPC produces the best results on the classification task ($p<.01$ for all metrics compared to all other methods), 
indicating the benefits of a Bayesian approach over SVM and BiLSTM.
However, unlike GPPL, GPC cannot be used to rank the arguments.
The results also show that PL+SVR does not reach the same performance as GPPL, 
suggesting that GPPL may benefit from the Bayesian integration of a GP with the preference likelihood. 

\subsection{Experiment 4: Conflicting and Noisy Data}

\begin{table}
\small
  \begin{tabularx}{\columnwidth}{ | l | X | X | X | X | X |}\hline
             & SVM & Bi-LSTM &GPPL medi.        &PL+ SVR     &GPC \\\hline
\multicolumn{6}{| l |}{Classification} \\   \hline             
Acc          & .70 & .73 & \textbf{.77}        &.75       &.73 \\
AUC          & .81 & .81 & .84        &.82       & \textbf{.86} \\
CEE          & .58 & .55 & \textbf{.50}     &.55       &.53 \\\hline
\multicolumn{6}{| l |}{Ranking} \\   \hline             
Pears.       & .32 & .22 & \textbf{.35}        &.31       & - \\
Spear.       & .43 & .30 & .54        & \textbf{.55}       & - \\
Kend.        & .31 & .21 & \textbf{.40}        & \textbf{.40}       & - \\
\hline
  \end{tabularx}
  \caption{Performance comparison on UKPConvArgCrowdSample using ling+GloVe features.}
  \label{tab:noisy}
\end{table}
We use UKPConvArgCrowdSample to introduce noisy data
and conflicting pairwise labels
to both the classification and regression tasks, to test
the hypothesis that GPPL would best handle unreliable crowdsourced data.
The evaluation uses gold labels from UKPConvArgStrict and UKPConvArgRank for the test set.
The results in Table \ref{tab:noisy} show that all methods perform worse compared to 
Experiment 3 due to the presence of errors in the pairwise labels. 
Here, GPPL produces the best classification accuracy and cross-entropy error (significant with $p\ll.01$ compared to all other methods except accuracy compared to GP+SVR, for which $p=.045$), while GPC has the highest AUC ($p\ll.01$ compared to all except GP+SVR, which was not significant). 
Compared to UKPConvArgStrict, the classification performance of GPC, SVM and BiLSTM decreased more than that of GPPL.
These methods lack a mechanism to resolve conflicts in the preference graph, unlike GPPL and PL+SVR, which handle conflicts through the preference likelihood.  
PL+SVR again performs worse than GPPL on classification metrics, although its ranking performance is comparable. 
For ranking, GPPL again outperforms SVM and BiLSTM in all metrics (significant with $p\ll.01$ in all cases except for SVM with Pearson's correlation).

\subsection{Experiment 5: Active Learning}

In this experiment, we hypothesized that GPPL provides more meaningful confidence estimates than SVM or BiLSTM,
which can be used to facilitate active learning in scenarios where labeled training data is expensive
or initially unavailable.
To test this hypothesis, we simulate an active learning scenario, in which an agent 
iteratively learns a model for each fold. Initially, $2$ pairs are chosen at random, then used to train the classifier. The agent then performs \emph{uncertainty sampling}~\cite{settles2010active} 
to select the $2$ pairs with the least confident classifications. 
The labels for these pairs are then added to the training set and 
used to re-train the model. We repeated the process until $400$ labels had been sampled. 

\begin{figure}
\centering
\includegraphics[width=0.9\columnwidth,trim=13 15 10 22.5,clip=true]{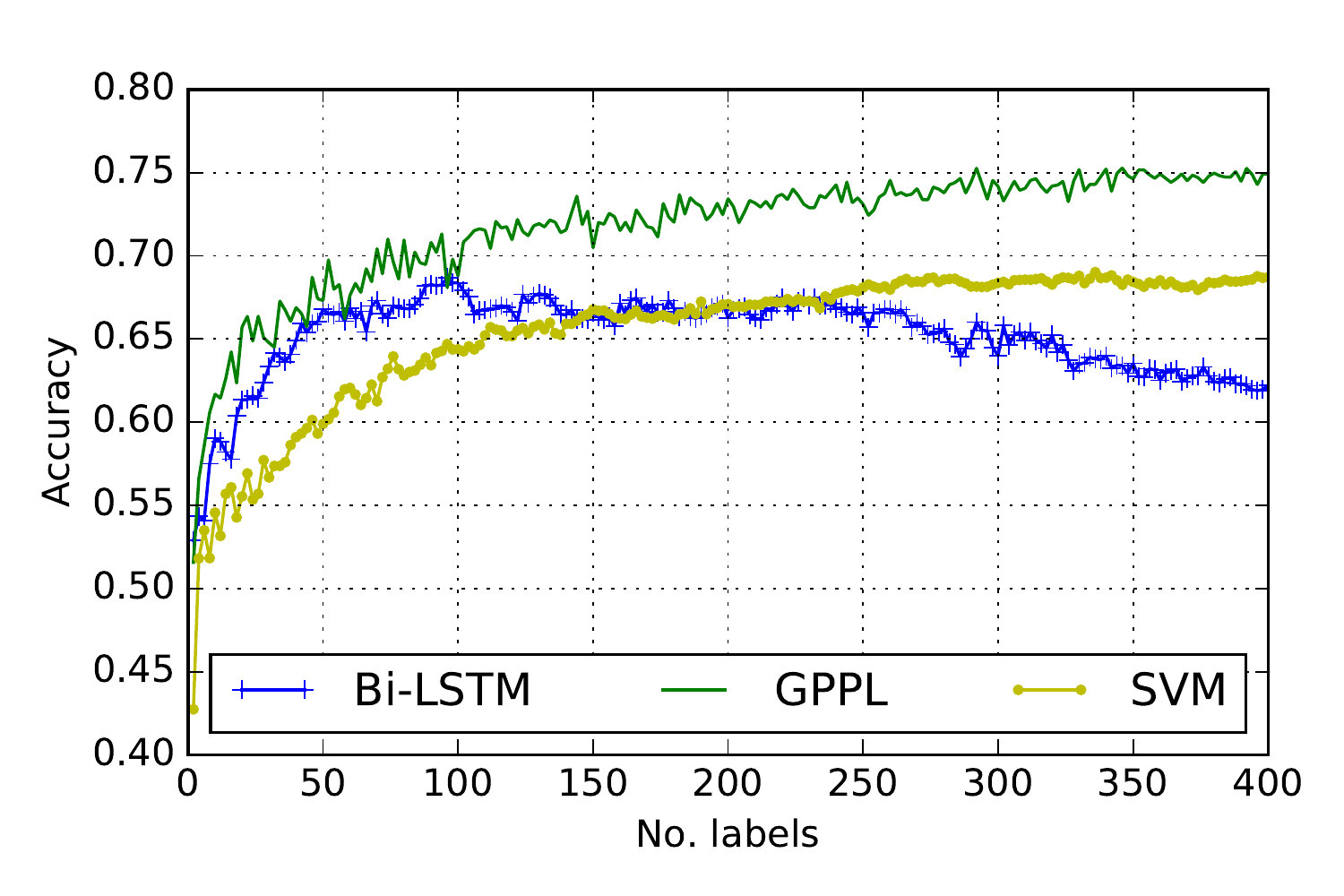}
\caption{Active learning simulation showing mean accuracy of preference pair classifications over 32 runs.}
\label{fig:active_learning}
\end{figure}

The result in Figure \ref{fig:active_learning} shows that GPPL
reaches a mean accuracy of 70\% with only 100 labels, while SVM and BiLSTM do not reach the same performance given 400 labels. 
After 100 labels, the 
performance of BiLSTM decreases. It has previously been shown ~\cite{cawley2011baseline,guyon2011results,settles2010active} that uncertainty sampling sometimes causes accuracy to decrease. If the model overfits to a small dataset, 
it can mis-classify some data points with high confidence so that they are not selected and corrected by uncertainty sampling.  
The larger number of parameters in the BiLSTM may make it may more prone to overfitting with small datasets than SVM or GPPL. 
The Bayesian approach of GPPL aims to further 
reduce overfitting by accounting for parameter uncertainty.
The results suggest that GPPL may be more suitable than the alternatives in cold-start scenarios with small amounts of labeled data. 

\subsection{Relevant Feature Determination}

\begin{figure}[h]
\includegraphics[width=\columnwidth, clip=True, trim=32 0 57 0]{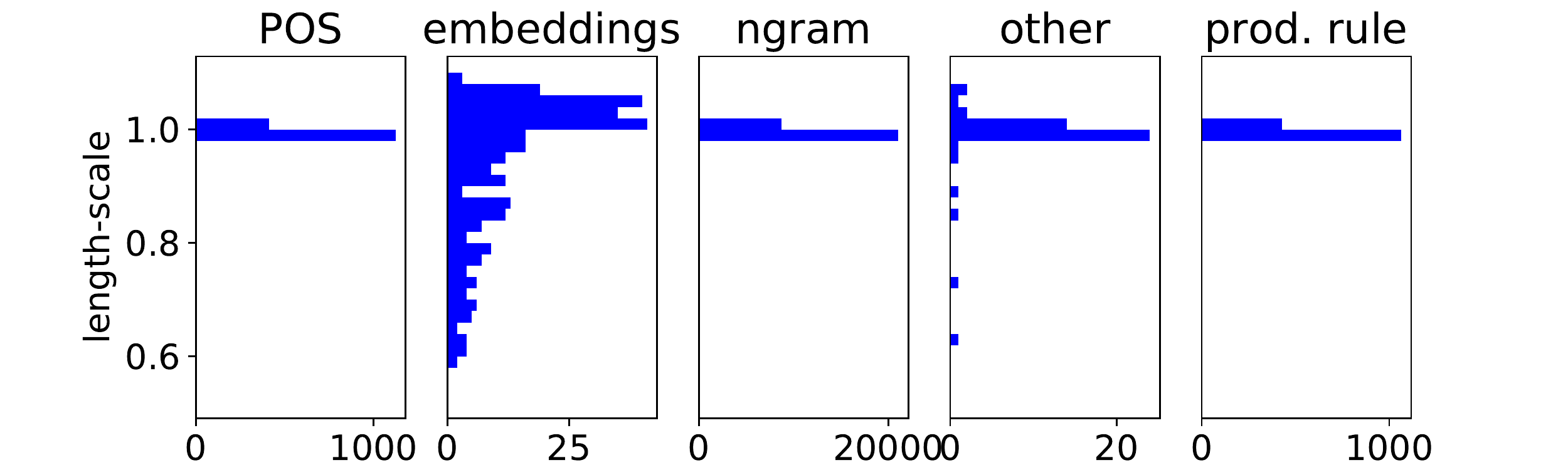}
\caption{Histograms of mean normalized length-scales on folds where MLII improved performance $>3\%$.}
\label{fig:boxplot}
\end{figure}

We now examine the length-scales learned by optimizing GPPL using MLII  
to identify informative features. 
A larger length-scale causes greater smoothing, 
implying that the feature is less relevant when predicting the convincingness function
than a feature with a small length-scale. 
Figure \ref{fig:boxplot} shows the distribution of length-scales for each category of
\emph{ling+GloVe} features, averaged over the folds in UKPConvArgStrict where MLII
optimization improved accuracy by $\>3\%$. The length-scales
were normalized by dividing by their median heuristic values, 
which were their initial values before optimization.
The widest distributions of length-scales are for the mean word embeddings and the ``other'' category.
A very large number of features have length-scales close to $1$,
which may mean that they are weakly informative, as their length-scales have not been increased,
or that there was insufficient data or time to learn their length-scales.
To limit computation time, the optimization algorithm was restricted to $25$ iterations, 
so may only have fully optimized features with larger gradients, 
leaving other features with normalized length-scales close to $1$.
 
Table \ref{tab:extreme_features} shows features with length-scales $<0.99$,
of which there are two production rule features and $18$ POS-n-gram features,
suggesting that the latter may capture more relevant aspects of grammar for convincingness. 
For n-grams, the relationship to convincingness may be topic-specific, 
hence they are not identified as important when the model is trained on $31$ different topics. 
The fact that MLII did not substantially shorten the length-scales for n-grams and POS n-grams 
corresponds to previous results ~\cite{persing2017can}, which found these feature sets less informative than other argument-related feature sets.
 
Table \ref{tab:extreme_features} also presents a breakdown of the ``other'' features into sentiment, ratio, count and NER features. 
The shortest length-scales are for sentiment features, pointing to a possible link between 
argumentation quality and sentiment. However, ``VeryPositive'' was the feature with
the largest length-scale, either because the median was a poor heuristic in this case or
because the feature was uninformative, perhaps because sarcastic statements can be confused with highly positive sentiment.
The short length-scale for the ``words $>6$ letters'' ratio suggest that some surface features may be informative,
despite previous work \cite{wei2016post} finding a set of surface features less informative than other feature sets. 
In this case, longer words may relate to more sophisticated and convincing arguments. 
\begin{table}[t]
\small
  \begin{tabularx}{\columnwidth}{l | X | r }
  Category & Feature & Length-scale\\
  \hline
  production rule & S$\to$NP,VP,.,	& 0.977\nonumber\\  
  production rule & S$\to$NP,VP,	& 0.988\nonumber\\  
  \hline
  POS-ngram & V-ADJ	& 0.950	\nonumber\\
  POS-ngram & PUNC-NN	& 0.974 \nonumber\\  
  POS-ngram & PR-PUNC	& 0.977	\nonumber\\
  POS-ngram & PP-V-PR	& 0.981\nonumber\\
  POS-ngram & NN-V-ADV	& 0.981\nonumber\\    
  \hline
  n-gram & ``.''	& 0.981\nonumber\\
  n-gram & ``to'' 	& 0.989\nonumber\\
  n-gram & ``in''	& 0.990\nonumber\\  
  \hline
  sentiment & Positive	& 0.636 \nonumber\\
  sentiment & VeryNegative	 & 0.842 \nonumber\\
  sentiment & Neutral	& 0.900 \nonumber\\
  sentiment & Negative & 0.967 \nonumber\\    
\emph{sentiment} & \emph{VeryPositive} & \emph{3.001} \nonumber \\
\hline
  ratio & words $>$ 6 letters & 0.734 \nonumber\\
  ratio & SuperlativeAdj	& 0.943 \nonumber\\
  ratio & InterjectionRate	& 0.986 \nonumber\\
  ratio &	SuperlativeAdv	& 0.987 \nonumber\\
 \hline
  count & words $>$ 6 letters	& 0.983 \nonumber\\  
 \hline
  NER & Location & 0.990 \nonumber  
  \end{tabularx}
  \caption{Normalized length-scales for linguistic features learned using MLII. Shows mean values over folds with $>3\%$ improvement. Includes all values $<0.99$, except for POS n-grams (only smallest 5 of 18 shown).  }
  \label{tab:extreme_features}
\end{table}

\subsection{Error Analysis}

We compared the errors when using GPPL opt. with mean GloVe embeddings
and with linguistic features. We
manually inspected the $25$ arguments most frequently
mis-classified by GPPL \emph{ling} and correctly classified by GPPL \emph{GloVe}.
We found that GPPL \emph{ling} mistakenly marked several arguments 
as less convincing when they contained grammar and spelling errors but otherwise
made a logical point. 
In contrast, arguments that did not strongly take a side and did not contain 
language errors were often marked mistakenly as more convincing.

We also examined the $25$ arguments most frequently misclassified by GPPL \emph{GloVe} but not by GPPL \emph{ling}.
Of the arguments that GPPL \emph{GloVe} incorrectly marked as more convincing, 
$10$ contained multiple exclamation marks and all-caps sentences. 
Other failures were very short arguments and underrating arguments containing the term `rape'.
The analysis suggests that the different feature sets identify different aspects of convincingness.

To investigate the differences between our best approach, GPPL opt. \emph{ling + GloVe}, 
and the previous best performer, SVM \emph{ling}, 
we manually examined $40$ randomly chosen false classifications, where one of 
either  \emph{ling + GloVe} or SVM was correct and the other was incorrect. 
We found that both SVM and GPPL falsely classified arguments that were either very short or long and complex, suggesting deeper semantic or structural understanding of the argument may be required. However, SVM also made mistakes
where the arguments contained few verbs.

We also compared the rankings produced by GPPL opt. (ling+GloVe), 
and SVM on UKPConvArgRank by examining the 20 largest deviations from the 
gold standard rank for each method. Arguments underrated by SVM and not GPPL often 
contained exclamation marks or common spelling errors (likely due to unigram or bigram features).
GPPL underrated short arguments with the ngrams ``I think", ``why?", and
``don't know", which were used as part of a rhetorical question
rather than to state that the author was uncertain or uninformed.
These cases may not be distinguishable by a GP given only \emph{ling + GloVe} features.

An expected advantage of GPPL is that it provides more meaningful uncertainty estimates for tasks such as active learning. 
We examined whether erroneous classifications correspond to more uncertain predictions
with GPPL \emph{ling} and SVM \emph{ling}.
For UKPConvArgStrict, the mean Shannon entropy
of the pairwise predictions from GPPL 
was .129 for correct predictions and 2.443 for errors,
while for SVM, the mean Shannon entropy was  .188 for correct predictions and 
1.583 for incorrect.
With both methods, more uncertain (higher entropy) predictions correlate with more errors,
but the more extreme values for GPPL suggest that its output probabilities more 
accurately reflect uncertainty than those produced by the SVM.

\section{Conclusions and Future Work}

We presented a novel Bayesian approach to predicting argument convincingness from pairwise labels using
Gaussian process preference learning (GPPL).
Using recent advances in approximate inference, we developed a scalable algorithm for GPPL 
that is suitable for large NLP datasets.
Our experiments demonstrated that our method significantly outperforms the state-of-the-art
on a benchmark dataset for argument convincingness, 
particularly when noisy and conflicting pairwise labels are used in training.
Active learning experiments showed that GPPL is an effective model for cold-start situations 
and that the convincingness of Internet arguments can be predicted reasonably 
well given only a small number of samples.
The results also showed that linguistic features and word embeddings provide complementary information,
and that GPPL can be used to automatically identify relevant features.

Future work will evaluate our approach on other NLP tasks 
where reliable classifications may be difficult to obtain, 
such as learning to classify text from implicit user feedback~\cite{joachims2002optimizing}.
We also plan to investigate training the GP using absolute scores in combination with pairwise labels.


\section*{Acknowledgements}

This work has been supported by the German Federal Ministry of Education and Research (BMBF) 
under the promotional reference 01UG1416B (CEDIFOR).
It also received funding from the European Union's
Horizon 2020 research and innovation programme (H2020-EINFRA-2014-2) under grant agreement No. 654021 
(Open-MinTeD). It reflects only the authors’ views and the EU is 
not liable for any use that may be made of the information contained therein. 
We would like to thank the TACL editors and reviewers for their effort and the valuable feedback we received from them.

\bibliographystyle{acl2012}
\bibliography{simpson_pref_learning_for_convincingness}


\end{document}